\pdfoutput=1

\documentclass[11pt,dvipsnames]{article}

\usepackage[final]{acl}

\usepackage{times}
\usepackage{amsmath}
\usepackage{latexsym}
\usepackage{bbding}
\usepackage{xcolor}
\usepackage[thicklines]{cancel}
\usepackage{tikz}
\usepackage{pgfplots}
\usepackage{subcaption}
\usetikzlibrary{matrix}
\usepackage[most]{tcolorbox}
\tcbuselibrary{breakable}
\tcbuselibrary{skins}
\usepackage{tabularray}
\usepackage{pgf-pie}
\usepackage{tablefootnote}
\usepackage{ragged2e} 
\usepackage{listings}
\usepackage{booktabs,makecell, multirow, tabularx}
\usepackage{float}

\usepackage[T1]{fontenc}

\usepackage[utf8]{inputenc}
\usepackage{calligra}
\usepackage{microtype}

\usepackage{inconsolata}

\usepackage{graphicx}
\newcommand{\datasetname}{\texttt{HomeBench}}
\newcommand{\halfcheck}{
    \CheckmarkBold\hspace{-0.375cm}\XSolidBrush
}
\definecolor{mycolor1}{RGB}{197,90,17} 
\definecolor{mycolor2}{RGB}{0,112,192} 
\title{\datasetname: Evaluating LLMs in Smart Homes with Valid and Invalid Instructions Across Single and Multiple Devices}

\author{
 \textbf{Silin Li\textsuperscript{1}},
 \textbf{Yuhang Guo\textsuperscript{1}},
 \textbf{Jiashu Yao\textsuperscript{1}},
 \textbf{Zeming Liu\textsuperscript{2} \thanks{Corresponding author: Zeming Liu}},
 \textbf{Haifeng Wang\textsuperscript{3}}
\\
 \textsuperscript{1}School of Computer Science and Technology, Beijing Institute of Technology\\
 \textsuperscript{2}School of Computer Science and Engineering, Beihang University
 \textsuperscript{3}Baidu Inc. 
\\
 \small{\normalsize 
  \{lisilin,guoyuhang,yaojiashu\}@bit.edu.cn
   \normalsize 
   zmliu@buaa.edu.cn 
   wanghaifeng@baidu.com
 }
}
\begin{document}
\maketitle
\begin{abstract}
Large language models (LLMs) have the potential to revolutionize smart home assistants by enhancing their ability to accurately understand user needs and respond appropriately, which is extremely beneficial for building a smarter home environment. While recent studies have explored integrating LLMs into smart home systems, they primarily focus on handling straightforward, valid single-device operation instructions. However, real-world scenarios are far more complex and often involve users issuing invalid instructions or controlling multiple devices simultaneously. These have two main challenges: LLMs must accurately identify and rectify errors in user instructions and execute multiple user instructions perfectly. To address these challenges and advance the development of LLM-based smart home assistants, we introduce \datasetname, the first smart home dataset with valid and invalid instructions across single and multiple devices in this paper. We have experimental results on 13 distinct LLMs; e.g., GPT-4o achieves only a 0.0\% success rate in the scenario of invalid multi-device instructions, revealing that the existing state-of-the-art LLMs still cannot perform well in this situation even with the help of in-context learning, retrieval-augmented generation, and fine-tuning. Our code and dataset are publicly available at the link\footnote{https://github.com/BITHLP/HomeBench}.
\end{abstract}

\section{Introduction}

Integrating large language models (LLMs) into smart home assistants can not only help users control devices with simpler and fewer instructions but also enable automatic device control based on the user's habits and history actions, without the need for explicit instructions~\cite{sasha,sega,yin2024harmonyhomeagentresponsive}. For instance, \citet{sasha,sega} introduced the Sasha and Sega systems, which preliminarily validated the feasibility of integrating LLMs into smart home assistants. \citet{yin2024harmonyhomeagentresponsive} further explored the capability of LLMs in handling ambiguous and no-instruction scenarios for smart home control, providing additional evidence for the promising application of large language models in this field. These researches primarily focus on the operation of valid single-device instructions~\cite{sasha,sega,yin2024harmonyhomeagentresponsive}, and the range and number of devices that LLMs can control are quite limited in these researches.

\begin{figure}[t]
    \centering
    \includegraphics[width=\linewidth]{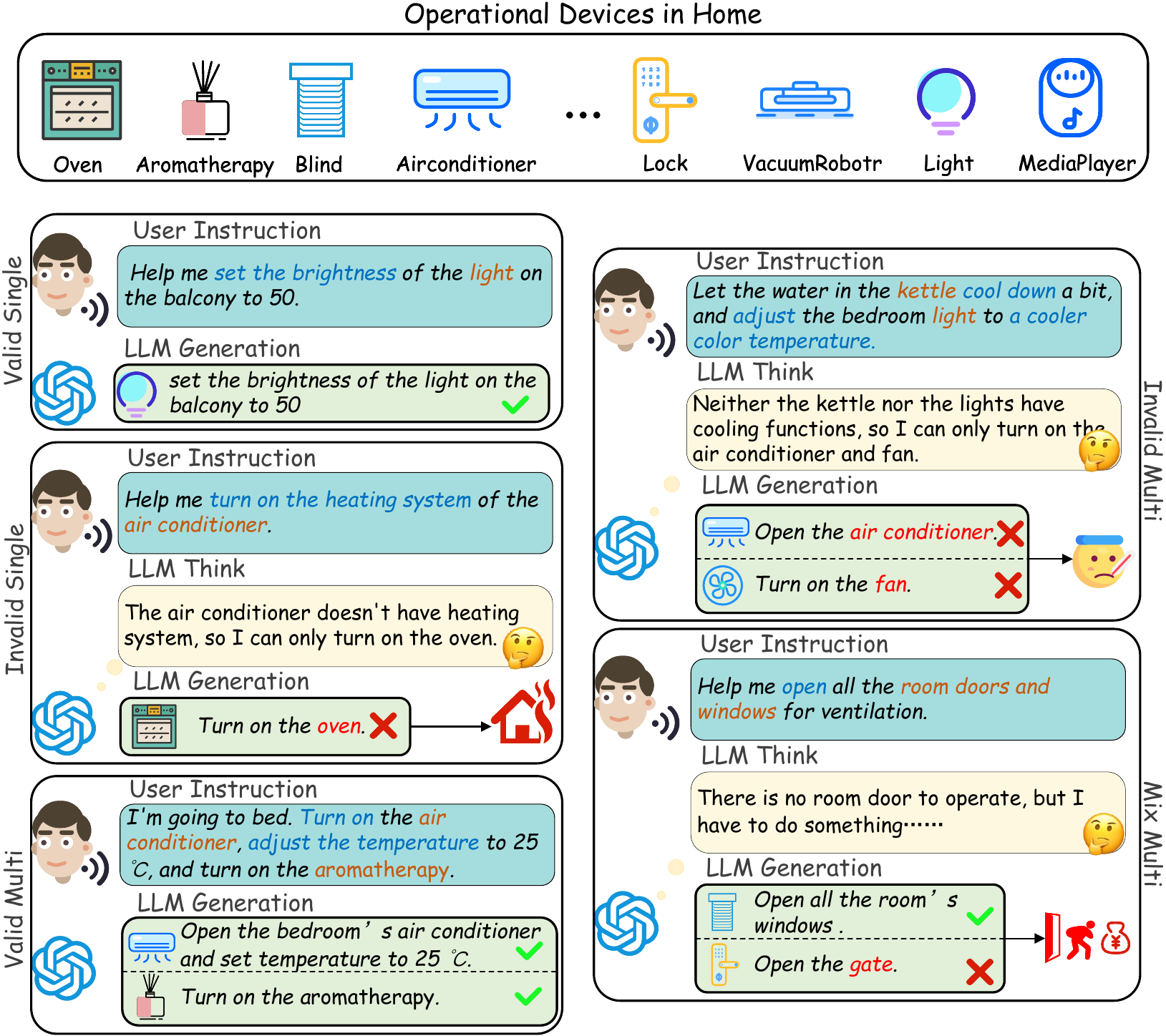}
    \caption{An example of valid single-device instruction, invalid single-device instruction, valid multi-device instruction, invalid multi-device instruction, and mixed multi-device instruction. \textcolor{mycolor1}{Orange} indicates the device to be operated, \textcolor{mycolor2}{blue} indicates the method of the device to be operated, and \textcolor{red}{red} indicates the wrong device to be operated. These mistake outputs are due to the operation of non-existent devices or non-existent device functions, causing the model to hallucinate and operate the wrong device.}
    \label{example}
\end{figure} 

However, in real-world scenarios, users may input invalid instructions for various reasons, such as being intoxicated or misspeaking. If the model faithfully executes these invalid instructions, it may negatively affect the user experience and safety. As Figure \ref{example} shows, mistakenly activating kitchen appliances or unlocking doors could pose significant safety risks. In addition, in real-world scenarios that involve multiple devices, users often need to control several devices simultaneously. Therefore, the development of LLMs capable of comprehensively managing multiple devices and understanding dynamic interactions in complex environments, while optimizing mechanisms for warning and correcting potential errors, becomes a key factor in the advancement of smart home systems.




To bridge these gaps, we introduced a new comprehensive evaluation benchmark: 
\datasetname, which is the first smart home dataset with valid and invalid instructions across single and multiple devices, with a diverse range and a large number of devices. Additionally, to facilitate flexible experimentation and scenario expansion, we developed a customizable virtual home environment that allows for the flexible setup of various devices and operational scenarios, offering a versatile experimental base for model testing and optimization. Specifically, we constructed 100 smart home scenarios covering more than 170k user operation instructions. Each scenario includes at least 47 smart devices across a minimum of 10 different types. For multi-device operation scenarios, we simulated complex interactions involving 1 to 10 devices to gauge the model's performance in highly intricate environments. By introducing \datasetname, we aim to pave the way for next-generation smart home assistants capable of seamlessly managing and coordinating numerous smart devices within modern IoT ecosystems.


Our main contributions can be summarized in the following three points:
\begin{itemize}
    \item 
    To the best of our knowledge, we are the first to study the model's ability in invalid instructions and multi-device operation instructions in smart homes.
    \item We proposed \datasetname, which contains valid and invalid instructions across single and multiple devices, covers complex operation scenarios, and developed a customizable virtual environment to support the generation and expansion of datasets.
    \item Our experimental results on 13 different LLMs show that almost all models perform poorly in this benchmark, especially on invalid instructions and multi-device instructions. Further analysis shows that simple in-context learning and fine-tuning significantly improve performance, but there is still a gap between them and practical application scenarios.
\end{itemize}

\section{Related Work}

\subsection{Home Assistant before LLM}
Smart home systems are built from interconnected IoT smart devices that monitor, sense, and control the home environment. Automating these devices not only improves quality of life and comfort but also optimizes resource efficiency~\cite{Ki2020CanAI}. Recent researches focus on enhancing the functionality of smart home systems through machine learning techniques. For example, \citet{9154007} develop a deep learning algorithm to identify user behavior using accelerometer data to achieve home automation. \citet{8261311} focus on voice-based home assistants, which are designed to understand user voice instructions. Using advanced NLP techniques, commercial solutions such as Bixby, Google Assistant, and Alexa provide user-friendly interfaces that can handle a variety of instructions and inquiries from shopping and setting reminders to device control and home automation. However, these modern home assistants still face challenges in handling ambiguous or complex instructions~\cite{10.1145/2858036.2858288}.

\subsection{Home Assistant on LLM}

In an attempt to overcome some of these challenges, recent work proposes using the reasoning capabilities of LLMs to better understand and carry out user instructions. ~\citet{shi2024bridginggapnaturaluser} propose AwareAuto, a system that standardizes user expressions and employs a two-step reasoning method powered by LLMs to generate automation solutions, achieving a 91.7\% accuracy rate in aligning with user intents. ~\citet{civitarese2024largelanguagemodelszeroshot} develope ADL-LLM, which transforms raw sensor data into textual descriptions to enable zero-shot recognition of daily living activities. GestureGPT \cite{1111111} integrates LLM reasoning capabilities into a triple-agent framework, enabling gesture analysis and execution of smart home tasks. Sasha~\cite{sasha} showcases the application of LLMs in smart home environments, demonstrating their ability to respond effectively to complex or ambiguous instructions. Sasha implements a decision-making process where each step (e.g., device selection or routine verification) is supported by LLMs. The SAGE system addresses the limitations of LLMs in lacking specific knowledge about users and households by dynamically constructing prompt trees and employing LLM-controlled discrete operation sequences. This significantly enhances its capabilities in information retrieval, user interaction, and device operation. However, these researches ignore the real scenarios of user invalid instructions and multiple operation instructions.

\section{Dataset Collection}

\begin{figure}[t]
    \centering
    \includegraphics[width=\linewidth]{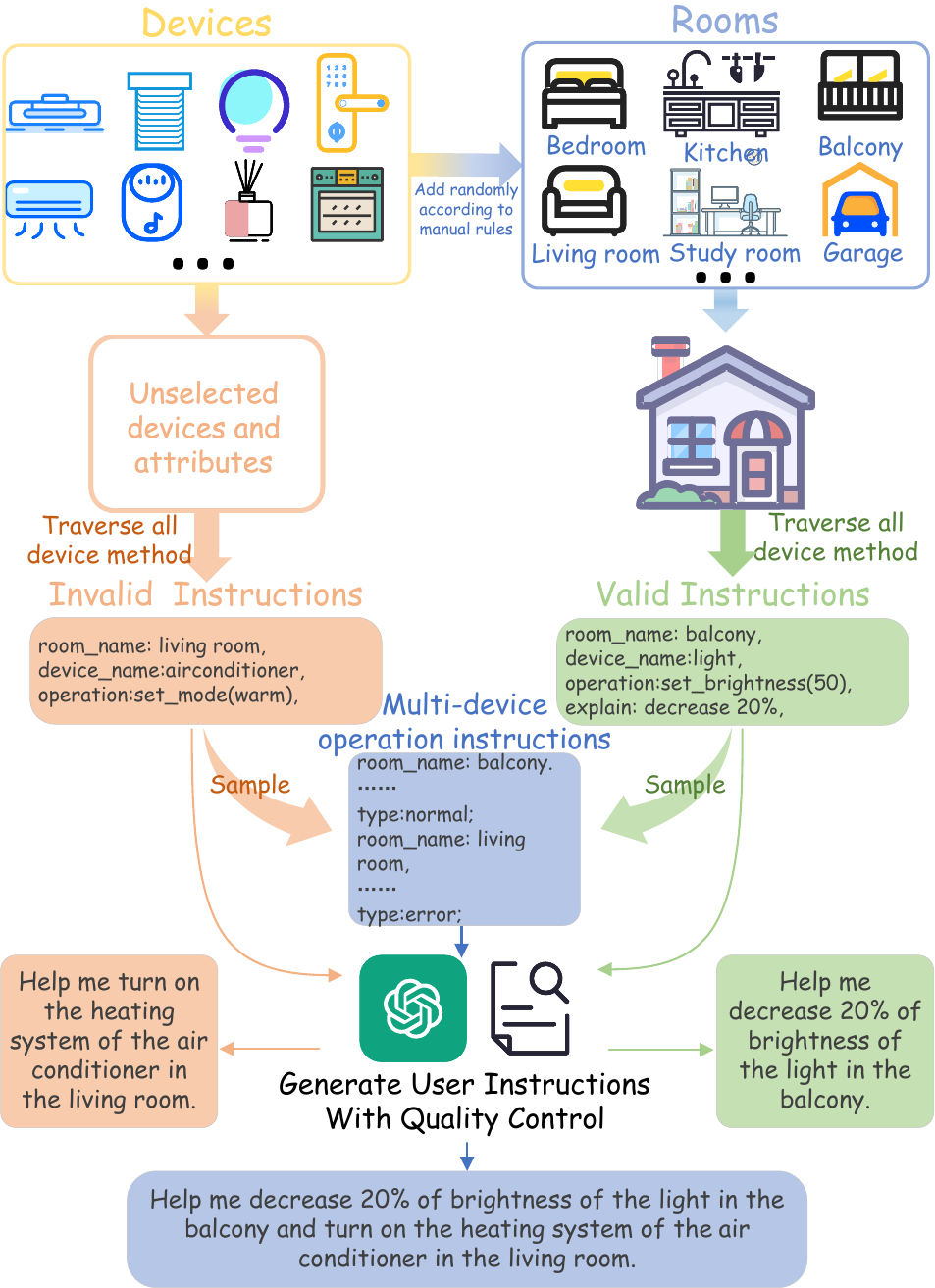}
    \caption{The whole process of collecting dataset \datasetname, from device construction, room setting, instructions generation, and user instructions synthesis to quality control.}
    \label{process}
\end{figure} 

\datasetname \quad aims to build high-quality smart home instruction call data to improve the single-device and multi-device call capabilities in complex smart home environments, focusing on the accurate response capabilities of LLM when facing user invalid instructions. In this section, we describe the dataset task format and then explain how to build the dataset efficiently and effectively. Figure \ref{process} shows the whole process of collecting dataset \datasetname.
\subsection{Task Definition}
Given a user instruction \( u \) and a virtual home environment \( H = \{r_1, r_2, r_3, \ldots\} \) with room information, each room \( r_i \) contains a set of operable devices \( D = \{d_1, d_2, \ldots\} \) in the current room. Each device \( d_j \) includes the state of its operable properties \( s \) and a set of callable methods \( M = \{m_1, m_2, m_3, \ldots\} \) and corresponding parameters. The LLM needs to decide which method of which device in which room to call to change the properties of the target device based on the user instruction \( u \), that is, \( h, u \to [r_i.d_j.m_k] \). If the user inputs an invalid instruction, the output is \( \text{error\_input} \).

\subsection{Data Categorization}

\begin{table}[ht]
\centering
\resizebox{\columnwidth}{!}{%
\begin{tabular}{l|llllll}
\hline
Dataset        & VS & IS & VM & IM & MM & SIZE \\ \hline
IFTTT~\cite{Yu_2021}          & \textcolor{yellow}{\halfcheck}   &  \textcolor{red}{\XSolidBrush}   &  \textcolor{yellow}{\halfcheck}  & \textcolor{red}{\XSolidBrush}   & \textcolor{red}{\XSolidBrush} & 50K+   \\
SEGA~\cite{sega}           &  \textcolor{green}{\Checkmark}  &  \textcolor{red}{\XSolidBrush}  & \textcolor{red}{\XSolidBrush}   &  \textcolor{red}{\XSolidBrush}  &  \textcolor{red}{\XSolidBrush} & 50  \\
Home Assistant\tablefootnote{https://huggingface.co/datasets/acon96/Home-Assistant-Requests} &  \textcolor{green}{\Checkmark}  &  \textcolor{red}{\XSolidBrush}  &  \textcolor{red}{\XSolidBrush}  &  \textcolor{red}{\XSolidBrush}  &  \textcolor{red}{\XSolidBrush}  & 30K+ \\ \hline
Ours           & \textcolor{green}{\Checkmark}   & \textcolor{green}{\Checkmark}   &  \textcolor{green}{\Checkmark}  &   \textcolor{green}{\Checkmark} &  \textcolor{green}{\Checkmark} & 170K+  \\ \hline
\end{tabular}%
}
\caption{Comparison with existing dataset at the instruction category for a fair comparison. "\textcolor{yellow}{\halfcheck}" represents the automated instruction generation for smart home operations, which differs significantly from our task setting. }
\label{tab1}
\end{table}

\begin{table*}[htp]
\centering
\begin{tabular}{cc|cccccc}
\hline
\multicolumn{2}{c|}{Statistics}                                & VS     & IS     & VM    & IM    & MM     & ALL     \\ \hline
\multicolumn{1}{c|}{\multirow{3}{*}{Train}} & Avg. Instruction Length & 11.13  & 11.27  & 32.48 & 27.82 & 69.37  & 25.59   \\
\multicolumn{1}{c|}{}                       & Avg. Number of Operating Device     & 1.00   & 1.00   & 2.91  & 2.51  & 6.26   & 2.30    \\
\multicolumn{1}{c|}{}                       & Number of Instructions             & 49,048 & 53,771 & 1,829 & 826   & 33,443 & 138,917 \\ \hline
\multicolumn{1}{c|}{\multirow{3}{*}{Valid}} & Avg. Instruction Length & 11.10  & 11.29  & 34.37 & 26.88 & 69.30  & 25.45   \\
\multicolumn{1}{c|}{}                       & Avg. Number of Operating Device      & 1.00   & 1.00   & 3.12  & 2.40  & 6.26   & 2.26    \\
\multicolumn{1}{c|}{}                       & Number of Instructions             & 6,145  & 6,731  & 232   & 124   & 4,132  & 17,364  \\ \hline
\multicolumn{1}{c|}{\multirow{3}{*}{Test}}  & Avg. Instruction Length & 11.19  & 11.28  & 33.49 & 25.88 & 68.81  & 25.13   \\
\multicolumn{1}{c|}{}                       & Avg. Number of Operating Device      & 1.00   & 1.00   & 2.98  & 2.30  & 6.21   & 2.29    \\
\multicolumn{1}{c|}{}                       & Number of Instructions             & 6,187  & 6,765  & 245   & 97    & 4,072  & 17,366  \\ \hline
\end{tabular}%

\caption{The data statistics of our proposed \datasetname.}
\label{tab2}
\end{table*}
Based on the types of instructions and the number of devices involved, we can categorize the data into five distinct types. Each type aligns with a typical use case in real-world scenarios, providing a comprehensive benchmark for assessing the capabilities of LLM in invoking smart home devices:
\begin{itemize}
    \item Valid single-device instruction (VS): The user issues an instruction that involves a single method of a single device. The instruction is correct and executable.
    \item Invalid single-device instruction (IS): The user issues an instruction that involves a single method of a single device. However, the instruction is incorrect and cannot be executed.
    \item Valid multi-device instruction (VM): The user issues instructions involving multiple methods across multiple devices. All instructions are correct and executable.
    \item Invalid multi-device instruction (IM): The user issues instructions involving multiple methods across multiple devices. However, all instructions are incorrect and cannot be executed.
    \item Mix multi-device instruction (MM): The user issues instructions involving multiple methods across multiple devices.  Some instructions are correct and executable, while others are incorrect and cannot be executed.
    
\end{itemize}


Table \ref{tab1} shows a detailed comparison of the dataset \datasetname ~with other popular datasets. Currently, most benchmarks focus on processing valid instructions of a single device, while common multi-device operation instructions and invalid instructions in real life are often ignored. This study makes important contributions in these two aspects.

\subsection{Data Collection}

Given the challenges in accessing smart home development environments on the market, we adopted a virtual smart home setting to reduce the entry barrier for our research. The dataset construction is divided into two primary phases: building virtual devices and environments and generating instructions.

For virtual device construction, we collected 15 commonly used smart home devices and identified the API methods that can be called through the smart home platform for each device. For example, devices such as robotic vacuum cleaners. We implemented these virtual devices using Python classes, allowing property modifications through API method calls. To account for performance differences among various brands and models, we simplified the methods for some devices, retaining only the essential smart home functions. For other devices, we randomly added additional methods to create a diverse pool of devices. To increase the complexity of the task, we designed twelve rooms, each with unique functionalities. Devices were randomly selected from the pool and assigned to rooms based on manually defined rules.

To ensure data quality while minimizing manual effort, we used GPT-3.5 to generate user instructions~\cite{wang2024surveydatasynthesisaugmentation}. For valid instructions, we first created all possible device operation commands within the virtual environment, specifying the room, device name, method called, and the method's input parameters. With the help of prompt rules and a few-shot examples, this information was provided to ChatGPT to synthesize user instructions.
For invalid instructions, we applied reverse rules to identify non-existent devices or missing methods within the virtual room and generated corresponding erroneous operation commands. These instructions were also processed through ChatGPT using the same approach.
For scenarios involving multiple instructions, we randomly selected commands from both the pool of valid operations and the invalid instruction pool. We then used the same synthesis method to generate user instructions.

To ensure the quality of the data, we sampled 10\% of the user instructions generated in each smart home scenario and 
conducted a rigorous evaluation from three dimensions:
fluency checks~\cite{wang-etal-2024-appbench,liu-etal-2020-towards-conversational} (language clarity), semantic alignment~\cite{iskander-etal-2024-quality} (instruction vs. device/room capabilities) and action consistency~\cite{iskander-etal-2024-quality} (synthesized instructions vs. executable machine instructions). Any mismatch resulted in immediate rejection (score=0). If an instruction scored more than 4, the data of the scenario was considered valid. This process was repeated until 100 high-quality smart home scenario data were collected.

\subsection{Data Statistic and Analysis}

Table \ref{tab2} shows the statistics of the dataset. Following previous works~\cite{zhang-etal-2023-fedlegal,kweon-etal-2023-open,cheng-etal-2022-iam}, we split the dataset in the conventional 8:1:1 ratio and provide detailed information on the average instruction length, number of instruction operation devices, and number of instructions according to the data types of VS, IS, VM, IM, and MM. 
Our data distribution reveals two notable patterns that warrant explanation: (1) the prevalence of single-device instructions over multi-device instructions, and (2) the relatively small number of IM instructions. Below we provide detailed explanations for these observations:
\begin{itemize}
    \item \textbf{Single-device instructions nums > multi-device instructions nums:}\quad  From the perspective of load balancing theory, humans tend to execute a single type of task multiple times rather than perform multiple distinct tasks simultaneously~\cite{SWELLER201137}. As a result, single-device operation commands are significantly more frequent than multi-device operation commands.
    \item \textbf{Small number of IM instructions:}\quad From a probabilistic standpoint, the likelihood of executing entirely valid or entirely invalid operations across multiple devices is much lower than that of mixed valid and invalid operations.
\end{itemize}

To demonstrate the complexity of our task, we statistic the types and numbers of devices in 100 homes. In 12 rooms with different functions, at least 47 devices can be operated, and at least 10 types of devices can be operated. More analysis can be found in the Appendix \ref{subsec:dc}.

\section{Experiment}
\subsection{Setting}
\textbf{Models}. 
Inspired by \citeauthor{fei-etal-2024-lawbench,huang-etal-2024-da}, we choose several LLMs from both open- and closed-source models, aiming to provide a comprehensive evaluation. Specifically, we choose Mistral-7B (Mistral-7B-v0.3)~\cite{jiang2023mistral7b}, LLaMa3-8b (llama3-8b-Instruct)~\cite{llama3modelcard}, the Qwen series (Qwen2.5-7B/32B/72B-Instruct)~\cite{qwen2.5}, the Gemma series (gemma-2-9b/27b-it)~\cite{gemmateam2024gemma2improvingopen} from opensource LLMs. Besides that, we also select GPT4 (GPT-4o) and deepseek v3~\cite{deepseekai2024deepseekv3technicalreport} from closed-source LLMs.
In addition, we also test reasoning-oriented models such as O1,O3 (o1-mini, o3-mini)~\cite{openai2025competitiveprogramminglargereasoning}, DeepSeek-R1~\cite{deepseekai2025deepseekr1incentivizingreasoningcapability}, and QWQ-32B\cite{qwq32b}.
We built two types of baseline models for experimental comparison: one is based on prompt, and the other is based on few-shot in-context learning~\cite{10.5555/3495724.3495883,zhou-etal-2024-enhancing-context}. As shown in Sec \ref{few-shot}, the performance improvement of the model after 3-shot is limited, so we chose 4-shot in the experiment.
In both types of baseline models, we embed the status information of each device in the virtual home and the methods that the device can call into the prompts to provide contextual information for the model. 
This setting is designed to ensure that the two methods are evaluated under the same information conditions to comprehensively compare their performance.
For specific details, please refer to Appendix \ref{subsec:es}.


\subsection{Evaluation Metrics}

In the experimental design of this study, we adhered to standardized testing paradigms in the field of embodied intelligence, employing a virtual environment as the test platform~\cite{zhu2024earbenchevaluatingphysicalrisk,zhang2024safeembodaisafetyframeworkmobile}. Based on the principle of functional equivalence, the virtual and physical devices maintain fully consistent testing criteria at the API interface level: a test is deemed passed when the model correctly generates API call sequences that comply with predefined specifications. Notably, the virtual testing environment offers unique experimental advantages, as it eliminates mechanical delays inherent to physical devices (by removing the need to wait for real-world feedback cycles), thereby significantly enhancing testing efficiency.

To evaluate the effectiveness of LLM in executing user instructions, we selected two key indicators: instruction execution success rate~\cite{shi2024bridginggapnaturaluser,sega} and F1 score.

\begin{table*}[htb]
\centering
\resizebox{\textwidth}{!}{%
    \begin{tabular}{ccccccccccccc}
    \hline
    \multirow{2}{*}{model} & \multicolumn{2}{c}{VS}         & \multicolumn{2}{c}{IS}         & \multicolumn{2}{c}{VM}         & \multicolumn{2}{c}{IM}         & \multicolumn{2}{c}{MM}         & \multicolumn{2}{c}{ALL}         \\ \cline{2-13} 
                           & SUCC           & F1             & SUCC           & F1             & SUCC           & F1             & SUCC           & F1             & SUCC           & F1             & SUCC           & F1             \\ \hline
    Mistral-7B             & 2.49           & 4.68           & 0.01           & 0.03           & 0.82           & 6.23           & 0.00           & 0.00           & 0.00           & 2.26           & 0.90           & 2.33           \\
    LLaMA3-8B              & 47.11          & 47.36          & 0.61           & 0.61           & 3.27           & 31.83          & 0.00           & 0.52           & 0.00           & 15.20          & 17.07          & 18.21          \\ \hline
    Qwen2.5-7B             & 30.40          & 31.92          & 0.00           & 0.00           & 13.69          & 33.81          & 0.00           & 0.00           & 0.00           & 16.13          & 11.02          & 16.24          \\
    Qwen2.5-32B            & 41.28          & 41.34          & 14.04          & 16.34          & 14.29          & 26.33          & 0.00           & 9.55           & 0.27           & 19.45          & 20.44          & 23.06          \\
    Qwen2.5-72B            & 42.90          & 42.91          & 11.47          & 12.57          & 19.59          & 38.06          & 1.03           & 9.07           & 0.15           & 24.79          & 20.07          & 25.92          \\ \hline
    Gemma2-9B              & 12.32          & 12.58          & 0.00           & 0.02           & 0.41           & 11.44          & 0.00           & 0.98           & 0.00           & 6.40           & 4.39           & 6.42           \\
    Gemma2-27B             & 12.32          & 13.08          & 0.93           & 1.03           & 0.82           & 11.18          & 0.00           & 2.33           & 0.02           & 6.93           & 4.77           & 7.34           \\ \hline
    Deepseek-V3            & 71.08          & 71.08          & 39.05          & 38.40          & 45.31          & 69.14          & 0.00           & 11.11          & 0.50           & 28.15          & 41.19          & 38.81          \\
    GPT-4o                 & 46.74          & 46.76          & 79.07          & 78.90          & 13.88          & 26.74          & 0.00           & 55.19          & 2.09           & 28.43          & 48.14          & 42.70          \\ \hline
    QWQ-32B                & 63.39          & 63.60          & 7.35          & 11.05          & 6.95           & 15.88         & 12.35           & 30.00          & 3.54           & 32.20          & 26.53          & 35.30          \\
    Deepseek-R1            & 65.67          & 65.76          & 83.74          & 83.76          & 0.00           & 1.95           & 0.00           & 23.22          & 1.41           & 23.22          & 56.39          & 40.96          \\
    o1-mini                & 38.65          & 39.46          & 20.34          & 30.34          & 1.19           & 2.09           & 0.00           & 27.75          & 0.73           & 15.13          & 21.85          & 23.20          \\
    o3-mini                & 27.57          & 28.16          & 45.43          & 57.51          & 4.56           & 13.64          & 0.00           & 33.08          & 0.21           & 16.30          & 27.66          & 27.32          \\ \hline
    Mistral-7B-ICL         & 47.10          & 53.63          & 0.15           & 0.17           & 17.96          & 50.07          & 0.00           & 0.00           & 0.02           & 22.74          & 17.10          & 24.15          \\
    LLaMA3-8B-ICL          & 68.08          & 68.17          & 2.38           & 2.37           & 30.61          & 60.63          & 0.00           & 2.71           & 0.17           & 28.69          & 25.65          & 30.78          \\ \hline
    Qwen2.5-7B-ICL         & 66.96          & 66.89          & 14.07          & 14.08          & 37.55          & 69.18          & 0.00           & 5.82           & 0.49           & 33.23          & 29.98          & 35.72          \\
    Qwen2.5-32B-ICL        & 77.24          & 77.48          & 69.33          & 69.27          & 45.31          & 73.46          & 3.09           & 44.44          & 5.38           & 52.71          & 56.44          & 59.84          \\
    Qwen2.5-72B-ICL        & 82.27          & 82.07          & 35.77          & 35.78          & 51.02          & 76.05          & 6.19           & 27.89          & 2.95           & 48.42          & 44.60          & 52.01          \\ \hline
    Gemma2-9B-ICL          & 59.72          & 59.65          & 21.67          & 21.74          & 33.47          & 62.54          & 3.09           & 20.62          & 1.55           & 42.17          & 30.57          & 41.67          \\
    Gemma2-27B-ICL         & 79.44          & 79.50          & 11.06          & 11.07          & 48.16          & 74.71          & 1.03           & 11.21          & 0.81           & 41.97          & 33.48          & 43.00          \\ \hline
    Deepseek-V3-ICL        & 80.00          & 80.06          & 58.58          & 58.48          & 54.69          & 77.29          & 11.34          & 32.22          & 5.35           & 52.74          & 53.41          & 58.74          \\
    GPT-4o-ICL             & 74.25          & 74.30          & 87.10          & 87.09          & 45.71          & 71.48          & 61.86          & 81.18          & 23.35          & 78.98          & 66.85          & 79.58          \\ \hline
    QWQ-32B-ICL            & 67.71          & 64.50          & 79.32          & 79.30          & 35.51          & 61.97          & 65.98          & 79.39          & 18.71          & 70.02          & 60.28          & 70.50          \\
    Deepseek-R1-ICL        & 76.48          & 77.46          & 84.18          & 84.77          & 46.19          & 70.75          & 66.27          & 83.03          & 28.28          & 77.73          & 67.62          & 78.82           \\
    o1-mini-ICL            & 75.42          & 75.94          & 86.80          & 86.90          & 53.39          & 75.52          & \textbf{79.17} & \textbf{89.09} & 32.25          & 81.37          & 69.47          & 81.42          \\
    o3-mini-ICL            & \textbf{83.77} & \textbf{84.25} & \textbf{88.36} & \textbf{88.38} & \textbf{57.51} & \textbf{80.49} & 64.04          & 85.19          & \textbf{38.49} & \textbf{85.57} & \textbf{74.44} & \textbf{85.75}          \\ \hline
    \end{tabular}%
    }
\caption{The main results of different LLMs on \datasetname. \textbf{Bold} indicates the best performance.}
\label{tab4}
\end{table*}

\textbf{Success Rate (Succ)}: We follow ~\citeauthor{shi2024bridginggapnaturaluser,sega}, this metric is an indicator to measure whether all device operations in a user instruction are executed successfully.

\textbf{F1}~\cite{li-etal-2024-hypergraph,cai-etal-2024-difinet,wang-etal-2024-order}: We first calculate the precision P, which is defined as the number of operations correctly predicted by the model divided by the total number of operations predicted by the model.
\begin{equation}
    Precision=\frac{\text { operation\_correct\_num }}{\text {operation\_pred\_num }}
\end{equation}
We calculate the recall R, which is the number of operations correctly predicted by the model divided by the total number of operations actually required in the user instruction.
\begin{equation}
    Recall=\frac{\text { operation\_correct\_num }}{\text {operation\_gold\_num }}
\end{equation}

The F1 score is 2*PR / (P+R), as usual.

\subsection{Results}

Table \ref{tab4} shows the results of different LLMs for different types of user instructions on \datasetname, respectively. Several conclusions can be drawn from the results.





\textbf{\textit{Overall, o3-mini achieves the best overall performance, while o1-mini outperforms o3-mini in certain situations, especially when processing only invalid instructions.}} Generally, with the exception of Deepseek-V3, gpt4o and reasoning models, other models significantly lag behind o3-mini in handling various types of instructions. This phenomenon can be attributed to the core difficulty of our task, which involves long-context attention. The model is required to thoroughly interpret a complete set of device states and methods to identify operable devices and those that are nonexistent. Among the models evaluated, o3-mini exhibits a markedly superior ability to manage long-range context.
Despite significant progress in LLMs, existing models still struggle with complex planning tasks, particularly when they involve hybrid instructions that require multi-device operations. In fact, even with the inclusion of ICL in MM tasks, the o3-mini success rate is only 38.49\%.

\textbf{\textit{As the size of the models increases, we observe improvements in performance in different types of instructions.}} This enhancement is particularly evident in the Qwen2.5 and Gemma2 series of models. Generally, larger models tend to deliver superior performance. However, the performance of most models significantly decreases in the presence of input invalid instructions or multi-device instructions, with some models getting 0\% at SUCC on specific tasks, such as IM tasks, indicating ongoing challenges in fully recognizing errors. Even advanced models like Deepseek-v3 can only partially detect errors within the instructions. Furthermore, these issues persist even when ICL is incorporated into the task setup, with most models struggling to overcome the challenges presented by IM tasks.

\textbf{\textit{With the introduction of ICL, the model demonstrated significant performance improvements in handling various instruction tasks.}} This phenomenon has been validated in all tested models, and some models have even achieved breakthrough performance gains. For example, the Mistral model demonstrated a more than 10-fold improvement in IS tasks, while the Gemma2-9b model saw its IS task success rate jump from 0\% to 21.67\%. Even for more complex tasks such as IM and MM, performance improvements were substantial. Notably, the Gemma2-9b model achieved a 20-point increase in these tasks. However, despite these significant enhancements, a considerable gap remains between the model's performance and the requirements of real-world applications, highlighting the need for further optimization and refinement.

\textbf{\textit{The performance of the models is heavily influenced by the complexity of the instructions.}} Analyzing performance in different scenarios reveals the following trend: VS>VM>IS>MM>IM. This trend is consistent across a majority of large language models, including Qwen2.5, Gemma2, LLaMA3-8B, Deepseek-v3, and Mistral-7B. It aligns with our expectations, as IM is the most complex task, followed by MM and IS, while VM is relatively simpler, and VS is the easiest. However, this trend does not hold for Deepseek-V3, GPT-4o and reasoning models. The key reason lies in the nature of IS and IM tasks, which require strong long-context attention. In these tasks, the model must track the states and methods of all devices to determine which operations are valid and which devices are unavailable. Deepseek-V3, GPT-4o, and reasoning models outperform other open-source models in terms of context handling capabilities~\cite{grattafiori2024llama3herdmodels,wang2024multimodalneedlehaystackbenchmarking,zhang-etal-2024-bench}.

\section{Analysis}

In this section, we conduct a comprehensive analysis, aiming to answer three research questions. RQ1: How does the few-shot influence the model performance? (Sec \ref{few-shot}) RQ2: Is it necessary to identify the room and device first to reduce the context window? (Sec \ref{rag}) and RQ3: What is the major bottleneck of current LLMs in our task (Sec \ref{error}), and can fine-tuning alleviate it? (Sec \ref{ft}).

\subsection{The Effects of Different ICL}
\label{few-shot}
\begin{figure}[htb]
\centering
\resizebox{\columnwidth}{!}{
$
\begin{matrix}
\begin{tikzpicture}
\begin{axis}[
legend columns=2,
xlabel={Shot Numbers},
    xmin=0, xmax=5,
    ymin=0, ymax=100,
    xtick={0,1,2,3,4,5},
    ytick={0,20,40,60,80,100},
    title={SUCC},
    legend style={ 
        legend pos =north east,legend columns=-1,}
]
\addplot[
    color=blue,
    mark=square,
    ] coordinates{
    (0,11.02)
    (1,26.98)
    (2,31.46)
    (3,30.41)
    (4,31.73)
    (5,31.43)
    };
    \addlegendentry{ALL}
\addplot[
    color=red,
    mark=star,
    ] coordinates{
    (0,30.40)
    (1,73.35)
    (2,66.85)
    (3,68.35)
    (4,67.30)
    (5,65.46)
    };
    \addlegendentry{VS}
\addplot[
    color=green,
    mark=diamond,
    ] coordinates{
    (0,0)
    (1,0.62)
    (2,18.21)
    (3,14.12)
    (4,18.33)
    (5,19.22)
    };
    \addlegendentry{IS}
\addplot[
    color=purple,
    mark=triangle,
    ] coordinates{
    (0,13.69)
    (1,43.27)
    (2,37.95)
    (3,39.18)
    (4,36.73)
    (5,36.33)
    };
    \addlegendentry{VM}
\addplot[
    color=orange,
    mark=+,
    ] coordinates{
    (0,0)
    (1,30.75)
    (2,0.07)
    (3,0.02)
    (4,0.39)
    (5,0.42)
    };
    \addlegendentry{MM}
\addplot[
    color=yellow,
    mark=*,
    ] coordinates{
    (0,0)
    (1,0)
    (2,0)
    (3,0)
    (4,0)
    (5,2.06)
    };
    \addlegendentry{IM}
\end{axis}
\end{tikzpicture}
&
\begin{tikzpicture}
\begin{axis}[legend columns=2,
xlabel={Shot Numbers},
    xmin=0, xmax=5,
    ymin=0, ymax=100,
    xtick={0,1,2,3,4,5},
    ytick={0,20,40,60,80,100},
    title={F1},
    legend style={ 
        legend pos =north east,legend columns=-1,}
]
\addplot[
    color=blue,
    mark=square,
    ] coordinates{
    (0,16.24)
    (1,32.63)
    (2,35.53)
    (3,35.00)
    (4,36.57)
    (5,36.52)
    };
    \addlegendentry{ALL}
\addplot[
    color=red,
    mark=star,
    ] coordinates{
    (0,31.92)
    (1,73.37)
    (2,66.83)
    (3,68.27)
    (4,67.24)
    (5,65.38)
    };
    \addlegendentry{VS}
\addplot[
    color=green,
    mark=diamond,
    ] coordinates{
    (0,0)
    (1,0.62)
    (2,18.17)
    (3,14.07)
    (4,18.30)
    (5,19.15)
    };
    \addlegendentry{IS}
\addplot[
    color=purple,
    mark=triangle,
    ] coordinates{
    (0,33.81)
    (1,71.38)
    (2,66.67)
    (3,69.93)
    (4,68.77)
    (5,68.22)
    };
    \addlegendentry{VM}
\addplot[
    color=orange,
    mark=+,
    ] coordinates{
    (0,16.13)
    (1,30.31)
    (2,31.82)
    (3,31.68)
    (4,33.31)
    (5,33.40)
    };
    \addlegendentry{MM}
\addplot[
    color=yellow,
    mark=*,
    ] coordinates{
    (0,0)
    (1,0)
    (2,8.35)
    (3,6.39)
    (4,6.27)
    (5,12.67)
    };
    \addlegendentry{IM}
\end{axis}
\end{tikzpicture}
\end{matrix}
$
}
\caption{The performance of Qwen2.5\-7B\-Instruct model in adding different types of data samples (the order of adding is: VS, IS, VM, MM, IM).}
\label{f-fewshot}
\end{figure}

Figure \ref{f-fewshot} shows the performance in the context of Qwen2.5-7B-Instruct when using different shots at the demonstrations, the performance of other models is in the Appendix\ref{subsec:es}. Since our data types are relatively rich, the types of shots added each time are also different. The specific order of addition is: VS, IS, VM, MM, IM. The shots here are obtained by random sampling in the training set. For detailed sampling methods and information, please refer to the description in the Appendix\ref{subsec:es}. The experimental results show that when we add different types of shots, the overall effect of the model is improved to a certain extent, and the corresponding data types also show different degrees of improvement. For example, when adding VS (1-shot), the success rate of the model on VS is increased from 30.40\% to 73.35\%; and when adding IM (5-shot), the F1 score of the model at IM increases from 6.27\% to 12.67\%. These results show that appropriately increasing the number and type of shots can help improve the performance of the model on specific datasets. 
However, we also observed that as the number of shots increases, performance declines in some tasks, particularly IS and VS. This may be because the provided shots primarily focus on user instructions and device operations while lacking information about the current status of home devices. As a result, the model struggles to accurately process these tasks, leading to performance degradation. Therefore, while increasing the number of shots generally improves overall performance, it is crucial to ensure the completeness and relevance of the shot content. Failing to include key contextual information may introduce information bias, negatively impacting the model's effectiveness in specific tasks.

\subsection{The Effects of RAG}\label{rag}
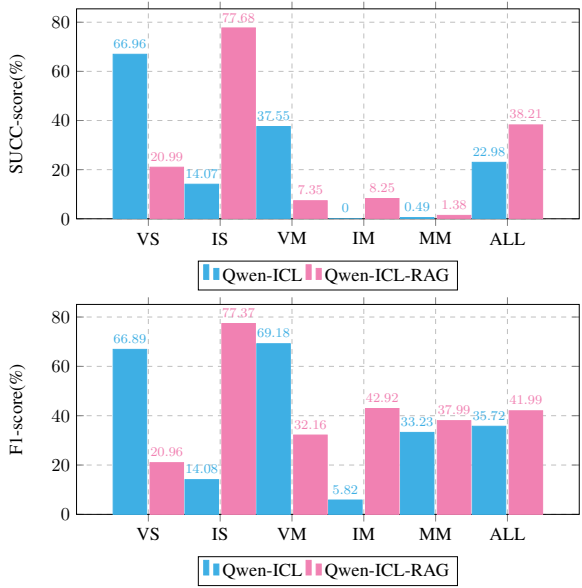
\begin{figure}[h]
    \centering
    \resizebox{\columnwidth}{!}{
    $
    \begin{matrix}
    
    \begin{tikzpicture}
    \begin{axis}
    [ybar , 
    grid=major,major grid style={dashed}, 
    ymin=0,  
    ylabel=SUCC-score(\%),  
    bar width=.7cm, 
    width=10cm,
    height=6cm,  
    symbolic x coords={VS,IS,VM,IM,MM,ALL},  
    x=1.5cm, 
    xtick=data, 
    nodes near coords, 
    nodes near coords style={font=\fontsize{8}{12}\selectfont}, 
    enlarge x limits=0.2, 
    legend style={at={(0.5,-0.2)},anchor=north,legend columns=-1}, 
    ] 
    \addplot[fill=CornflowerBlue,color=CornflowerBlue] coordinates {(VS, 66.96) (IS, 14.07) (VM, 37.55) (MM, 0.49) (IM,0.0) (ALL,22.98)};
    \addplot[fill=CarnationPink,color=CarnationPink] coordinates {(VS, 20.99) (IS, 77.68) (VM, 7.35) (MM, 1.38) (IM,8.25) (ALL,38.21)};
    \legend{Qwen-ICL,Qwen-ICL-RAG}
    \end{axis} 
    \end{tikzpicture}
    \\
    \begin{tikzpicture}
    \begin{axis}
    [ybar , 
    grid=major,major grid style={dashed}, 
    ymin=0,  
    ylabel=F1-score(\%),  
    bar width=.7cm, 
    width=10cm,
    height=6cm,  
    symbolic x coords={VS,IS,VM,IM,MM,ALL},  
    x=1.5cm, 
    xtick=data, 
    nodes near coords, 
    nodes near coords style={font=\fontsize{8}{12}\selectfont}, 
    enlarge x limits=0.2, 
    legend style={at={(0.5,-0.2)},anchor=north,legend columns=-1}, 
    ] 
    \addplot[fill=CornflowerBlue,color=CornflowerBlue] coordinates {(VS, 66.89) (IS, 14.08) (VM, 69.18) (MM, 33.23) (IM,5.82) (ALL,35.72)};
    \addplot[fill=CarnationPink,color=CarnationPink] coordinates {(VS, 20.96) (IS, 77.37) (VM, 32.16) (MM, 37.99) (IM,42.92) (ALL,41.99)};
    \legend{Qwen-ICL,Qwen-ICL-RAG}
    \end{axis} 
    \end{tikzpicture}
    \end{matrix}
    $
    }
    \caption{The performance gap between Qwen-ICL and Qwen-ICL-RAG.}
    \label{fig:RAG}
\end{figure}
In our task setup, it is necessary to provide the status information of smart devices and the callable methods in the prompt, which results in a high token count, exceeding 3k tokens. However, each operation typically involves only 1 to 10 devices, making much of the information redundant. To reduce redundancy and improve overall performance, we attempted to optimize using the retrieval-augmented generation (RAG)~\cite{NEURIPS2020_6b493230,wang-etal-2024-rag,shi-etal-2024-generate}. We segmented the context based on room information, retrieving the most relevant rooms for the task at hand and providing them to the LLM. Specific details of the setup can be found in the Appendix \ref{subsec:es}. The results are shown in Figure \ref{fig:RAG}. It can be observed that the experimental results were anomalous: tasks that the model was originally proficient in showed a decline in performance, while tasks it was less proficient exhibited significant improvement. We conducted an error analysis of the results and found that the retrieved rooms contained considerable errors. Although the context was shortened, the amount of useful information was also reduced, leading to a decline in the performance of tasks involving valid instructions, while the performance of tasks involving invalid instructions improved. 
We have added a statistical analysis of retrieval context errors in Table \ref{retrieval context errors} and output cases when retrieved incorrect context in the Appendix \ref{sec:appendix}.
\begin{table}[!h]
\resizebox{\columnwidth}{!}{%
\begin{tabular}{c|c|c}
\hline
                                                                      & Single Device Instruction & Multi Devices Instructions \\ \hline
\begin{tabular}[c]{@{}c@{}}None Useful \\ State Context\end{tabular}  & 74.49\%                   & 46.47\%                    \\ \hline
\begin{tabular}[c]{@{}c@{}}None Useful \\ Method Context\end{tabular} & 75.97\%                   & 48.74\%                    \\ \hline
\begin{tabular}[c]{@{}c@{}}Lack Useful \\ State Context\end{tabular}  & -                         & 46.65\%                    \\ \hline
\begin{tabular}[c]{@{}c@{}}Lack Useful \\ Method Context\end{tabular} & -                         & 45.04\%                    \\ \hline
\end{tabular}%
}
\caption{Statistics on Errors in RAG Context Retrieval.
We conducted a statistical analysis to determine whether the context retrieved using embedding vector similarity for the Qwen model aligns with the required context for the given input instructions.}
\label{retrieval context errors}
\end{table}

\subsection{Error Analysis}\label{error}

\begin{table}[!ht]
\centering
\resizebox{\columnwidth}{!}{%
\begin{tabular}{c|lcc}
\hline
Error Type                                                                             & \multicolumn{1}{c}{Input}                                                                                                                                                                                                                    & Generated and Golden       & Ratio                                                                                                        \\ \hline
Unfaithfulness                                                                         & \begin{tabular}[c]{@{}l@{}}...\\ master\_bedroom:\\ \quad light:\\    \qquad state: off\\ ...\\ \textless{}User instruction:\textgreater\\ Please turn the light in \\ the master bedroom to 50\%.\end{tabular}                                        & \begin{tabular}[c]{@{}c@{}}\textcolor{BrickRed}{error\_input}\\ \textcolor{OliveGreen}{master\_bedroom.light.}\\ \textcolor{OliveGreen}{set\_brightness(50)}\end{tabular}             & 27.93                         \\ \hline
\multirow{2}{*}{\begin{tabular}[c]{@{}c@{}}\\ \\ \\ \\In-context \\ Attention Error\end{tabular}} & \begin{tabular}[c]{@{}l@{}}...\\ living\_room:\\ \quad media\_player:\\   \qquad  state: on\\   \qquad  volume:28\\ ...\\ \textless{}User instruction:\textgreater\\ Decrease the volume of \\ the media player on the \\ balcony by 3 percent.\end{tabular} & \begin{tabular}[c]{@{}c@{}}\textcolor{BrickRed}{living\_room.media\_}\\ \textcolor{BrickRed}{player.set\_volume(20)}\\ \textcolor{OliveGreen}{living\_room.media\_}\\ \textcolor{OliveGreen}{player.set\_volume(25)}\end{tabular}  & \multirow{2}{*}[-12.5ex]{43.39}   \\ \cline{2-3} 
                                                                                       & \begin{tabular}[c]{@{}l@{}}...\\ store\_room:\\  (There is no airpurifiers)\\ ...\\ \textless{}User instruction:\textgreater\\ Help me jump the air \\ purifier in the storage\\ room to strong mode.\end{tabular}                          & \begin{tabular}[c]{@{}c@{}}\textcolor{BrickRed}{store\_room.airpur}\\ \textcolor{BrickRed}{ifiers.set\_mode(high)}\\ \textcolor{OliveGreen}{error\_input}\end{tabular}                                        \\ \hline
Key Error                                                                                                                      
                                                                                        & \begin{tabular}[c]{@{}l@{}}...\\ guest\_room:\\\quad  heating:\\     \qquad state:on\\   \qquad  mode: heating\\ ...\\ \textless{}User instruction:\textgreater\\ Set the heating in the guest\\ bedroom to fan only mode.\end{tabular}                       & \begin{tabular}[c]{@{}c@{}}\textcolor{BrickRed}{guest\_bedroom.heating.}\\ \textcolor{BrickRed}{set\_mode("fan\_only")}\\ \textcolor{OliveGreen}{guest\_bedroom.heating}\\ \textcolor{OliveGreen}{.set\_mode(fan\_only)}\end{tabular} & 34.39\\      \hline
\end{tabular}%
}
\caption{The three main error types of GPT-4o test results are unfaithfulness, in-context
attention error and key error, respectively. \textcolor{BrickRed}{Red} represents the generated \textcolor{OliveGreen}{green} represents the golden answer. Since multi-device operations generate multiple instructions, a user instruction may contain multiple error types. We also provide the proportion of various error types for various data types in the Appendix \ref{subsec:es}.}
\label{error analysis}
\end{table}

We performed an error analysis on the experimental results to identify potential bottlenecks in the current best model, GPT-4o. As shown in Table \ref{error analysis}, the errors can be categorized into the following three types:

1) Unfaithfulness: The model fails to execute user instructions accurately, incorrectly judging a correct instruction as incorrect. For example, in the case shown in the table, the model mistakenly assumes that a light is inoperable simply because the master bedroom light is turned off, leading to an incorrect response.

2) In-Context Attention Error: The model failed to notice key information contained in the context, leading to the output of invalid instructions. For instance, the user instruction in the table was to adjust the relative volume, but the model did not accurately grasp the current device's status information. Another issue is that the model did not consider the status and methods of all devices comprehensively, leading to an inability to confirm which devices were operable or non-existent.

3) Key Error: The model's output does not adhere to the prompt’s specified format, preventing effective statistical analysis of the results.

\subsection{The Effects of Fine-tuning}
\label{ft}

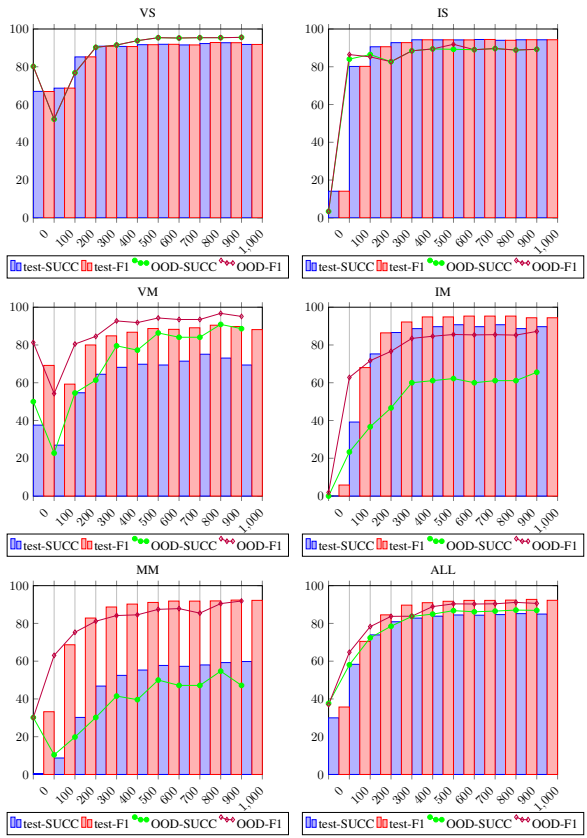
\begin{figure}[htb]
\centering
\resizebox{\columnwidth}{!}{
$
\begin{matrix}
\begin{tikzpicture} 
\begin{axis}[ legend style={at={(0.5,-0.2)},anchor=north,legend columns=-1}, ybar interval, 
    xmin=0, xmax=1100,
    ymin=0, ymax=100,
    xtick={0,100,200,300,400,500,600,700,800,900,1000,1100},
    ytick={0,20,40,60,80,100},
    title={VS},
    x tick label style={
    rotate=45
    }
] 
\addplot coordinates {(0,66.96)(100,68.66) (200,85.21) (300,90.88) (400,90.77) (500,91.69) (600,91.90) (700,91.56) (800,92.29) (900,92.73) (1000,91.81)(1100,0)}; 
\addplot coordinates {(0,66.89)(100,68.69) (200,85.24) (300,90.85)(400,90.70) (500,91.68) (600,91.89) (700,91.56) (800,92.84) (900,92.74) (1000,91.81)(1100,0)}; 
\addplot[sharp plot,mark=*, color=green] coordinates {(0,80.18)(100,52.21) (200,76.81) (300,90.27)(400,91.50) (500,93.81) (600,95.39) (700,95.22) (800,95.39) (900,95.39) (1000,95.57)};
\addplot[sharp plot,mark=diamond, color=purple] coordinates {(0,80.18)(100,52.29) (200,76.75) (300,90.27)(400,91.50) (500,93.81) (600,95.39) (700,95.22) (800,95.39) (900,95.39) (1000,95.57)};
\legend{test-SUCC,test-F1,OOD-SUCC,OOD-F1} 
\end{axis} 
\end{tikzpicture}
&
\begin{tikzpicture} 
\begin{axis}[  legend style={at={(0.5,-0.2)},anchor=north,legend columns=-1}, ybar interval, 
    xmin=0, xmax=1100,
    ymin=0, ymax=100,
    xtick={0,100,200,300,400,500,600,700,800,900,1000,1100},
    ytick={0,20,40,60,80,100},
    title={IS},
    x tick label style={
    rotate=45
    }
] 
\addplot coordinates {(0,14.07)(100,80.16) (200,90.58) (300,92.76) (400,94.28) (500,94.25) (600,94.22) (700,94.44) (800,94.01) (900,94.29) (1000,94.29)(1100,0)}; 
\addplot coordinates {(0,14.08)(100,80.20) (200,90.60) (300,92.78)(400,94.28) (500,94.27) (600,94.22) (700,94.45) (800,94.02) (900,94.31) (1000,94.30)(1100,0)}; 
\addplot[sharp plot,mark=*, color=green] coordinates {(0,3.39)(100,84.06) (200,86.45) (300,82.67)(400,88.44) (500,89.44) (600,89.24) (700,89.04) (800,89.64) (900,88.84) (1000,89.24)};
\addplot[sharp plot,mark=diamond, color=purple] coordinates {(0,3.39)(100,86.45) (200,85.24) (300,82.67)(400,88.44) (500,89.44) (600,91.89) (700,89.04) (800,89.64) (900,88.84) (1000,89.24)};
\legend{test-SUCC,test-F1,OOD-SUCC,OOD-F1} 
\end{axis} 
\end{tikzpicture}
\\
\begin{tikzpicture} 
\begin{axis}[  legend style={at={(0.5,-0.2)},anchor=north,legend columns=-1}, ybar interval, 
    xmin=0, xmax=1100,
    ymin=0, ymax=100,
    xtick={0,100,200,300,400,500,600,700,800,900,1000,1100},
    ytick={0,20,40,60,80,100},
    title={VM},
    x tick label style={
    rotate=45
    }
] 
\addplot coordinates {(0,37.55)(100,26.94) (200,54.69) (300,64.49) (400,68.16) (500,69.80) (600,69.39) (700,71.43) (800,75.10) (900,73.06) (1000,69.39)(1100,0)}; 
\addplot coordinates {(0,69.18)(100,59.26) (200,80.00) (300,84.83)(400,86.75) (500,88.69) (600,88.28) (700,89
.10) (800,90.47) (900,89.79) (1000,88.14)(1100,0)}; 
\addplot[sharp plot,mark=*, color=green] coordinates {(0,50)(100,22.73) (200,54.55) (300,61.36)(400,79.54) (500,77.27) (600,86.36) (700,84.09) (800,84.09) (900,90.90) (1000,88.63)};
\addplot[sharp plot,mark=diamond, color=purple] coordinates {(0,81.30)(100,54.25) (200,80.49) (300,84.55)(400,92.74) (500,91.86) (600,94.30) (700,93.49) (800,93.50) (900,96.74) (1000,95.12)};
\legend{test-SUCC,test-F1,OOD-SUCC,OOD-F1} 
\end{axis} 
\end{tikzpicture}
&
\begin{tikzpicture} 
\begin{axis}[  legend style={at={(0.5,-0.2)},anchor=north,legend columns=-1}, ybar interval, 
    xmin=0, xmax=1100,
    ymin=0, ymax=100,
    xtick={0,100,200,300,400,500,600,700,800,900,1000,1100},
    ytick={0,20,40,60,80,100},
    title={IM},
    x tick label style={
    rotate=45
    }
] 
\addplot coordinates {(0,0)(100,39.18) (200,75.26) (300,86.60) (400,88.66) (500,89.69) (600,90.72) (700,89.69) (800,90.72) (900,88.66) (1000,89.69)(1100,0)}; 
\addplot coordinates {(0,5.82)(100,68.02) (200,86.41) (300,92.17)(400,94.85) (500,94.85) (600,95.30) (700,95.30) (800,95.30) (900,94.41) (1000,94.41)(1100,0)}; 
\addplot[sharp plot,mark=*, color=green] coordinates {(0,0)(100,23.33) (200,36.67) (300,46.66)(400,60) (500,61.11) (600,62.22) (700,60) (800,61.11) (900,61.11) (1000,65.55)};
\addplot[sharp plot,mark=diamond, color=purple] coordinates {(0,1.85)(100,62.80) (200,71.68) (300,76.64)(400,83.48) (500,84.62) (600,85.55) (700,85.34) (800,85.45) (900,85.23) (1000,87.19)};
\legend{test-SUCC,test-F1,OOD-SUCC,OOD-F1} 
\end{axis} 
\end{tikzpicture}
\\
\begin{tikzpicture} 
\begin{axis}[  legend style={at={(0.5,-0.2)},anchor=north,legend columns=-1}, ybar interval, 
    xmin=0, xmax=1100,
    ymin=0, ymax=100,
    xtick={0,100,200,300,400,500,600,700,800,900,1000,1100},
    ytick={0,20,40,60,80,100},
    title={MM},
    x tick label style={
    rotate=45
    }
] 
\addplot coordinates {(0,0.49)(100,8.72) (200,30.21) (300,46.81) (400,52.48) (500,55.30) (600,57.74) (700,57.27) (800,58.01) (900,59.26) (1000,59.82)(1100,0)}; 
\addplot coordinates {(0,33.23)(100,68.69) (200,82.84) (300,88.69)(400,90.23) (500,91.11) (600,91.83) (700,91.81) (800,91.92) (900,92.35) (1000,92.21)(1100,0)}; 
\addplot[sharp plot,mark=*, color=green] coordinates {(0,30.16)(100,10.38) (200,19.81) (300,30.18)(400,41.51) (500,39.62) (600,50) (700,47.16) (800,47.16) (900,54.71) (1000,47.17)};
\addplot[sharp plot,mark=diamond, color=purple] coordinates {(0,30.16)(100,63.06) (200,75.30) (300,81.15)(400,84.17) (500,84.55) (600,87.45) (700,87.84) (800,85.45) (900,90.49) (1000,91.87)};
\legend{test-SUCC,test-F1,OOD-SUCC,OOD-F1} 
\end{axis} 
\end{tikzpicture}
&
\begin{tikzpicture} 
\begin{axis}[  legend style={at={(0.5,-0.2)},anchor=north,legend columns=-1}, ybar interval,
    xmin=0, xmax=1100,
    ymin=0, ymax=100,
    xtick={0,100,200,300,400,500,600,700,800,900,1000,1100},
    ytick={0,20,40,60,80,100},
    title={ALL},
    x tick label style={
    rotate=45
    }
] 
\addplot coordinates {(0,29.98)(100,58.33) (200,73.92) (300,80.88) (400,82.83) (500,83.84) (600,84.47) (700,84.35) (800,84.67) (900,85.19) (1000,84.95)(1100,0)}; 
\addplot coordinates {(0,35.72)(100,70.50) (200,84.53) (300,89.68)(400,90.97) (500,91.72) (600,92.20) (700,92.20) (800,92.34) (900,92.72) (1000,92.24)(1100,0)}; 
\addplot[sharp plot,mark=*, color=green] coordinates {(0,37.64)(100,58.07) (200,72.38) (300,78.50)(400,83.70) (500,84.93) (600,86.76) (700,86.15) (800,86.53) (900,87.06) (1000,86.91)};
\addplot[sharp plot,mark=diamond, color=purple] coordinates {(0,37.01)(100,64.73) (200,78.30) (300,83.74)(400,83.82) (500,88.92) (600,90.36) (700,90.28) (800,90.38) (900,91.14) (1000,90.59)};
\legend{test-SUCC,test-F1,OOD-SUCC,OOD-F1} 
\end{axis} 
\end{tikzpicture}

\end{matrix}
$
}
\caption{Performance of Qwen2.5-7B-Instruct in test dataset and OOD dataset under different training steps.}
\label{finetuning}
\end{figure}

We fine-tuned Qwen2.5-7B-Instruct on the full training dataset using LoRA for 2 epochs, testing every 100 training steps. The test results are shown in Figure \ref{finetuning}. As the training steps increased, the performance across all tasks improved significantly, but after 1 epoch of training, the performance improvement gradually plateaued. Ultimately, the model outperformed GPT-4o-ICL in all tasks. For most tasks, except VM and MM, both SUCC and F1 scores exceeded 80\%. However, despite training, the SUCC scores for VM and MM remained below 70\%. To further assess generalization, we conducted an Out-of-Distribution (OOD) evaluation, introducing two previously unseen devices. The results show that even on these new devices, the model’s performance remained consistent with its test set results, demonstrating the robustness and diversity of our dataset. For detailed information on the OOD data, please refer to Appendix \ref{subsec:es}. However, while fine-tuning led to performance improvements, it did not fundamentally resolve key issues. A deeper analysis revealed that the primary reasons for low SUCC scores were In-Context Attention Errors and Unfaithfulness, highlighting areas that still require further.

\section{Conclusion}
To advance the development of LLM-based smart home assistants and address the challenges posed by invalid instructions and multi-device instructions in real-world scenarios, we propose \datasetname, the first smart home dataset with valid and invalid instructions across single and multiple devices in this paper. We evaluated 13 LLMs and explored various methods to enhance their performance, including RAG, ICL, and fine-tuning. Our work lays a solid foundation for the optimization and application of LLM-based smart home assistants.

\section*{Limitations}
We acknowledge the following limitations of our work. Our dataset consists of only English monolingual data, whereas real-world applications may involve multiple languages. We did not account for the differences between devices of various brands during the construction process, although such differences do exist in real-world applications. We believe that these applications can typically be distinguished by modifying device names. Moreover, the choice of brand often depends on personal preference, which is beyond the scope of this study.

\section*{Ethical Statement}
During our research, we have thoroughly reviewed and ensured compliance with ethical standards. The equipment and rooms we constructed were carefully selected, and strict quality control was applied during the data synthesis process to avoid any form of offensive or biased content. Therefore, we believe there are no ethical issues with our research. The data used is ethically sourced, the analysis is fair, and all procedures adhere to established ethical guidelines.

\section*{Acknowledgement}
We thank all the anonymous reviewers for their insightful and valuable comments. This work is supported by the National Natural Science Foundation of China (Grant No. U21B2009, 62406015) and Beijing Institute of Technology Science and Technology Innovation Plan (Grant No. 23CX13027).
\bibliography{custom}

\appendix

\section{Appendix}
\label{sec:appendix}

\begin{table*}[]
\centering

\begin{tcolorbox}[colback=gray!10,
			colframe=black,
			width=14cm,
			arc=2mm, auto outer arc,
			title={Prompt}, breakable, enhanced jigsaw,
			before upper={\parindent15pt\noindent},	]
You are an artificial intelligence, skilled at converting machine instructions into commands issued by humans. Don't be polite in tone, give instructions directly, and use a colloquial expression format. If multiple machine instructions are given, you need to convert them into one sentence. If the instruction contains an "explain" field, please note to generate human instructions of the amplitude adjustment type. Here are some examples:

Machine instructions: \{'instruction': 'set\_intensity(30)', 'device': 'humidifier', 'explain': 'increase 2 percent', 'room': 'guest\_bedroom'\}

Human instructions: Increase the intensity of the humidifier by 2\% in the guest bedroom.

Machine instructions: \{'instruction': 'set\_temperature(26)', 'device': 'air\_conditioner', 'explain': 'decrease 1 percent', 'room': 'guest\_bedroom'\}

Human instructions: Lower the air conditioner temperature in the guest bedroom by 1 degree.

Here are the machine instructions you want to modify:

\end{tcolorbox}	
   
\caption{The prompt used to prompt LLM to generate the user instruction. In the prompt, we provide two examples to illustrate different adjustment methods more clearly. The first example demonstrates how to set a specific value directly for precise control of the target parameter. In contrast, the second example adopts a relative adjustment approach, modifying the value based on its current state to achieve dynamic tuning.}
\label{tabl5}
\end{table*}

\begin{figure}[!ht]
    \centering
    \includegraphics[width=\linewidth]{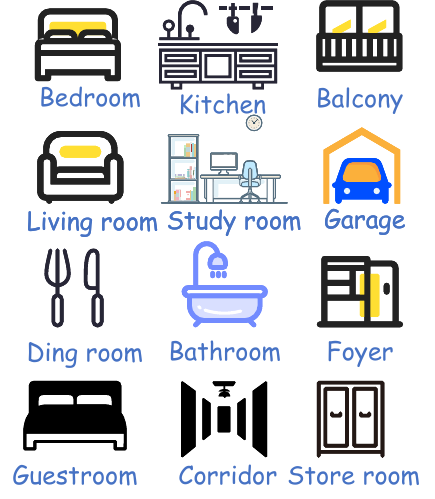}
    \caption{ALL room types in \datasetname.}
    \label{tabl4}
\end{figure}

\begin{table*}[]
\centering
\resizebox{\textwidth}{!}{%
\begin{tabular}{cc}
\hline
device type          & method                                                                  \\ \hline
LightDevice          & turn\_on;turn\_off;set\_brightness;set\_color                           \\
AirConditionerDevice & turn\_on;turn\_off;set\_temperature;set\_mode;set\_fan\_speed;set\_swing \\
HeatingDevice        & turn\_on;turn\_off;set\_temperature;set\_mode;set\_fan\_speed            \\
FanDevice            & turn\_on;turn\_off;set\_speed;set\_swing                                 \\
GarageDoorDevice     & open;close                                                              \\
BlindsDevice         & open;close                                                              \\
CurtainDevice        & open;close;set\_degree                                                  \\
AirPurifiersDevice   & turn\_on;turn\_off;set\_mode;set\_fan\_speed                            \\
WaterHeaterDevice    & turn\_on;turn\_off;set\_temperature;set\_mode                            \\
MediaPlayerDevice    & play;pause;stop;set\_volume;set\_song;set\_artist;set\_style            \\
VacuumRobotrDevice   & start,pause;stop;charge;set\_mode;set\_cleaning\_area                   \\
AromatherapyDevice   & turn\_on;turn\_off;set\_intensity;set\_interval                          \\
TrashDevice          & pack                                                                    \\
HumidifierDevice     & turn\_on;turn\_off;set\_intensity;set\_mode                              \\
DehumidifiersDevice  & turn\_on;turn\_off;set\_intensity;set\_mode                              \\ \hline
\end{tabular}%
}
\caption{All device types and their methods in \datasetname.}
\label{tabl41}
\end{table*}

\begin{figure*}
\centering
\begin{lstlisting}[
    numbers=left, % 显示行号
    numberstyle=\tiny, % 行号字体
    keywordstyle=\color{blue!70}, % 关键字颜色
    commentstyle=\color{red!50!green!50!blue!50}, % 注释颜色
    frame=shadowbox, % 为代码块添加阴影框
    rulesepcolor=\color{red!20!green!20!blue!20}, % 阴影框颜色
    escapeinside=``, % 允许在代码块中使用 LaTeX 命令
    xleftmargin=2em, xrightmargin=2em, aboveskip=1em, % 设置代码块的边距
    framexleftmargin=2em,% 阴影框左边距
    language=python,
    breaklines=true,
]
class VisualLivingRoom:
    def __init__(self) -> None:
        self.name = "living_room"
        self.devices = []
        self.unexist_devices = []
        self.devices.append(random.choice(LightDeviceList)("on"))
        if random.random() > 0.5:
            self.devices.append(random.choice(AirConditionerDeviceList)("on"))
            self.unexist_devices.append(random.choice(HeatingDeviceList)("on"))
            self.unexist_devices.append(random.choice(FanDeviceList)("on"))
        else:
            self.devices.append(random.choice(HeatingDeviceList)("on"))
            self.devices.append(random.choice(FanDeviceList)("on"))
            self.unexist_devices.append(random.choice(AirConditionerDeviceList)("on"))

        random.choice([self.devices, self.unexist_devices]).append(random.choice(CurtainDeviceList)("on"))
        random.choice([self.devices, self.unexist_devices]).append(random.choice(AirPurifiersDeviceList)("on"))

        if random.random() > 0.5:
            self.devices.append(random.choice(HumidifierDeviceList)("on"))
            self.unexist_devices.append(random.choice(DehumidifiersDeviceList)("on"))
        else:
            self.devices.append(random.choice(DehumidifiersDeviceList)("on"))
            self.unexist_devices.append(random.choice(HumidifierDeviceList)("on"))

        random.choice([self.devices, self.unexist_devices]).append(random.choice(AromatherapyDeviceList)("on"))
        random.choice([self.devices, self.unexist_devices]).append(random.choice(TrashDeviceList)("on"))
        random.choice([self.devices, self.unexist_devices]).append(random.choice(MediaPlayerDeviceList)("play"))
        random.choice([self.devices, self.unexist_devices]).append(random.choice(PetFeederDeviceList)("on"))

        self.random_initialize()
        self.state = self.get_status()

\end{lstlisting}
\caption{Smart home virtual room configuration: ensuring logical device selection.}
\label{fig:rule}
\end{figure*}


\begin{figure}
    \centering
    \resizebox{\columnwidth}{!}{
    \begin{tikzpicture}
    \begin{axis}
    [ybar , 
    grid=major,major grid style={dashed}, 
    ymin=0,  
    ylabel=Nums,  
    bar width=.5cm, 
    width=10cm,
    height=6cm,  
    symbolic x coords={Min,Max,Avg.},  
    x=1.5cm, 
    xtick=data, 
    nodes near coords, 
    nodes near coords style={font=\fontsize{8}{12}\selectfont}, 
    enlarge x limits=0.2, 
    legend style={at={(0.5,-0.2)},anchor=north,legend columns=-1}, 
    ] 
    \addplot[fill=CornflowerBlue,color=CornflowerBlue] coordinates {(Min, 47) (Max, 64) (Avg., 57.09)};
    \addplot[fill=CarnationPink,color=CarnationPink] coordinates {(Min, 10) (Max, 14) (Avg., 13.09)};
    \legend{Device Num,Device Type}
    \end{axis} 
    \end{tikzpicture}
    }
    \caption{The statistics of the number and types of devices in 100 home environments.}
    \label{tab3}
\end{figure}
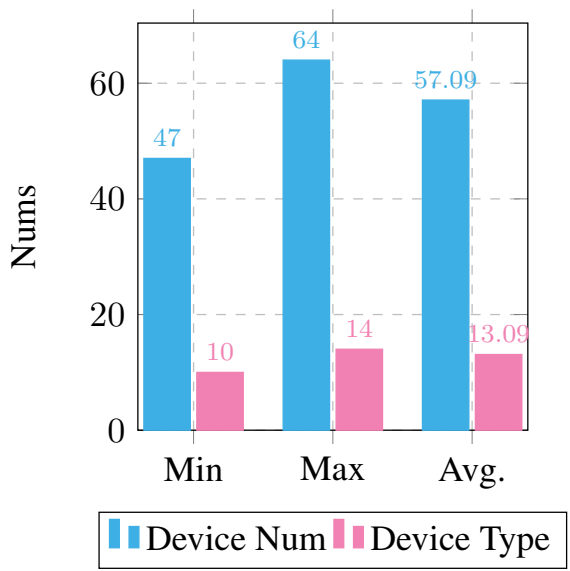


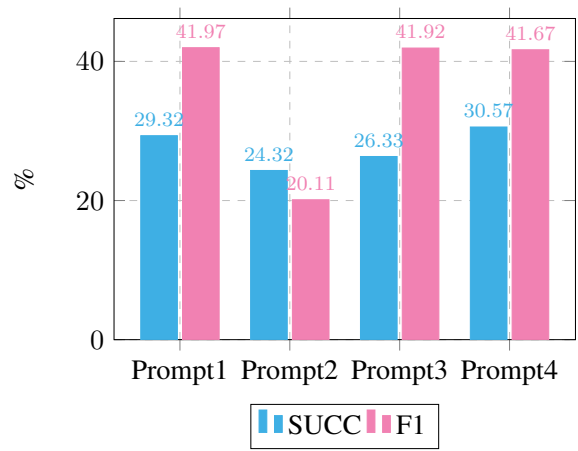
\begin{figure}
    \centering
    \resizebox{\columnwidth}{!}{
    \begin{tikzpicture}
    \begin{axis}
    [ybar , 
    grid=major,major grid style={dashed}, 
    ymin=0,  
    bar width=.5cm, 
    width=10cm,
    height=6cm,  
    ylabel=\%,
    symbolic x coords={Prompt1,Prompt2,Prompt3,Prompt4},  
    x=1.5cm, 
    xtick=data, 
    nodes near coords, 
    nodes near coords style={font=\fontsize{8}{12}\selectfont}, 
    enlarge x limits=0.2, 
    legend style={at={(0.5,-0.2)},anchor=north,legend columns=-1}, 
    ] 
    \addplot[fill=CornflowerBlue,color=CornflowerBlue] coordinates {(Prompt1,29.32) (Prompt2,24.32) (Prompt3,26.33)(Prompt4,30.57)};
    \addplot[fill=CarnationPink,color=CarnationPink] coordinates {(Prompt1,41.97) (Prompt2,20.11) (Prompt3,41.92)(Prompt4,41.67)};
    \legend{SUCC,F1}
    \end{axis} 
    \end{tikzpicture}
    }
    \caption{Performance of Gemma2-9B-Instruct on test dataset under different prompts. Prompt 1 and Prompt 2 were manually written. Prompt 3 was generated by modifying Prompt 1 using GPT-4. Prompt 4 was also manually written and is the final prompt we used.}
\label{tabl10}
\end{figure}

\begin{table*}[t]
    \centering
    \begin{tcolorbox}[colback=gray!10,
			colframe=black,
			width=14cm,
			arc=2mm, auto outer arc,
			title={Prompt},
			before upper={\parindent15pt\noindent},	]
            
You are 'Al', a helpful AI Assistant that controls the devices in a house. Complete the following task as instructed or answer the following question with the information provided only. The current status of the device and the methods it possesses are provided below, please only use the methods provided. Output \"error\_input\" when operating non-existent attributes and devices. Only output machine instructions and enclose them in \{\}.
    
    \textcolor{Cerulean}{<home\_state>  }
    
    \textcolor{Cerulean}{The following provides the status of all devices in each room of the current household, the adjustable attributes of each device, and the threshold values for adjustable attributes:}
    
    \textcolor{Cerulean}{......(Home device status, Table \ref{tabl7} shows an example.)}
    
    \textcolor{Cerulean}{</home\_state> }
    
    \textcolor{mycolor1}{<device\_method>}
    
    \textcolor{mycolor1}{The following provides the methods to control each device in the current household:}
    
    \textcolor{mycolor1}{......(Device methods, Table \ref{tabl8} show an example.)}
    
    \textcolor{mycolor1}{</device\_method>}
    
    -------------------------------
    
    Here are the user instructions you need to reply to.
    
    <User instructions:> 
    
    The user instructions.
    
    <Machine instructions:>

\end{tcolorbox}	
\caption{The prompt used to prompt LLM to generate an operation. \textcolor{Cerulean}{Blue} is home state and \textcolor{mycolor1}{orange} is device method.}
\label{tabl6}
\end{table*}

\begin{table*}[]
\centering
\begin{tcolorbox}[colback=gray!10,
			colframe=black,
			width=14cm,
			arc=2mm, auto outer arc,
			title={Example},
			before upper={\parindent15pt\noindent},	]
    master\_bedroom:  
    
    ---light     
    
                -------state: off

                -------brightness: 35 (range: 0 - 100)  
                
                -------color: [232, 50, 198]  
                
    --heating      
    
                -------state: off   
                
                -------mode: fan\_only (options['heat', 'fan\_only'])    
                
                -------fan\_speed: auto (options['auto', 'low', 'medium', 'high'])         
                
    --fan         
    
                ------state: on  
                
                ------swing: auto (options['auto', 'up', 'middle', 'down'])     

            .......

\end{tcolorbox}	
\caption{Examples of home device status.}
\label{tabl7}
\end{table*}

\begin{table*}[]
\centering
\begin{tcolorbox}[colback=gray!10,
			colframe=black,
			width=14cm,
			arc=2mm, auto outer arc,
			title={Example},]
    master\_bedroom.light.turn\_on();
    
    master\_bedroom.light.turn\_off();
    
    master\_bedroom.light.set\_brightness(brightness:int);
    
    master\_bedroom.light.set\_color(color:typing.Tuple[int, int, int]);
    
    master\_bedroom.heating.turn\_on();
    
    master\_bedroom.heating.turn\_off();
    
    master\_bedroom.heating.set\_mode(mode:str);
    
    master\_bedroom.heating.set\_fan\_speed(fan\_speed:str);

    master\_bedroom.fan.turn\_on();

    master\_bedroom.fan.turn\_off();
    
    master\_bedroom.fan.set\_swing(swing:str);

\end{tcolorbox}	
    \caption{Examples of home device methods.}
    \label{tabl8}
\end{table*}

\begin{table*}[]
\centering
\resizebox{\textwidth}{!}{%
\begin{tabular}{c|c|c}
\hline
Date Type & User Instruction                                                                                                                                                                                                                                                                                                                                                & Output                                                                                                                                                                                                 \\ \hline
VS        & Set the brightness of the light on the balcony to 50.                                                                                                                                                                                                                                                                                                           & balcony.light.set\_brightness(50)                                                                                                                                                                      \\ \hline
IS        & Pause the media player in the living room.                                                                                                                                                                                                                                                                                                                      & error\_input                                                                                                                                                                                           \\ \hline
VM        & \begin{tabular}[c]{@{}c@{}}Lower the air conditioner temperature in the guest bedroom to 20 degrees,\\ set the brightness of the light in the foyer to 50,\\ and increase the intensity of the dehumidifier in the study room by 32\%.\end{tabular}                                                                                                             & \begin{tabular}[c]{@{}c@{}}guest\_bedroom.air\_conditioner.set\_temperature(20),\\ foyer.light.set\_brightness(50),\\ study\_room.dehumidifiers.set\_intensity(70)\end{tabular}                        \\ \hline
IM        & \begin{tabular}[c]{@{}c@{}}Set the intensity of the humidifier to 60 in the study room, \\ adjust the volume of the media player to 60 on the balcony, \\ and set the degree of the curtain to 20 on the balcony.\end{tabular}                                                                                                                                  & \begin{tabular}[c]{@{}c@{}}error\_input,\\ error\_input,\\ error\_input,\end{tabular}                                                                                                                  \\ \hline
MM        & \begin{tabular}[c]{@{}c@{}}Set the air conditioner temperature to 16 degrees in the guest bedroom, \\ turn off the lights in the study room, \\ decrease the media player volume by 30 percent on the balcony, \\ set the dehumidifier intensity to 60 in the master bedroom, \\ and adjust the heating temperature to 27 degrees in the bathroom.\end{tabular} & \begin{tabular}[c]{@{}c@{}}guest\_bedroom.air\_conditioner.set\_temperature(16),\\ study\_room.light.turn\_off(),\\ error\_input,\\ error\_input,\\ bathroom.heating.set\_temperature(27)\end{tabular} \\ \hline
\end{tabular}%
}
\caption{Examples of different data types in \datasetname.}
\label{dataset case}
\end{table*}

\subsection{Data Collection}
\label{subsec:dc}

Table \ref{tabl5} presents the prompts we generated for user instructions using ChatGPT-3.5. To enhance the quality and accuracy of the generated instructions, we incorporated two example cases into the prompt. These examples provide a reference for the model, helping it better understand the task requirements and generate more precise and contextually appropriate responses to user instructions.

Figure \ref{tabl4} illustrates the multiple rooms with distinct functions that we designed. Each room's functionality has been carefully planned to ensure it aligns with realistic usage scenarios, covering a wide range of smart home applications. This structured design allows the model to handle diverse real-world contexts more effectively.

Figure \ref{fig:rule} shows a virtual room construction example, we demonstrate how to select and configure smart home devices based on predefined rules to ensure consistency and logical compatibility. As shown in the code, certain devices in a living room setting have mutual exclusion constraints. For instance, an air conditioner, heater, and fan cannot coexist in the same space to prevent functional conflicts. Similarly, a humidifier and a dehumidifier are mutually exclusive to maintain proper environmental control. For other device categories, such as lighting, television, and audio systems, we randomly select devices from a predefined list and determine whether to add them to the current room. If the room already contains devices, new selections may be added as an expansion; if the room is empty, a new set of devices may be initialized, creating diverse configurations of virtual rooms. This construction logic not only ensures a reasonable selection of devices but also lays a foundation for further smart home scenario simulations and optimizations.

Table \ref{tabl41} lists all device categories included in our dataset, along with the corresponding methods that each category may contain. To enhance device diversity and ensure compatibility with the multifunctional nature of various smart home brands, we assigned a set of basic functions (e.g., turning on/off) to each device and randomly added additional functions. This approach allows devices to adapt dynamically to different situations, providing a more flexible and realistic user experience.

Figure \ref{tab3} presents statistical insights into the distribution and types of devices across 100 virtual home environments. Within 12 functionally distinct rooms, users can interact with at least 47 operable devices spanning 10 different device categories. This comprehensive setup ensures a diverse and practical testing environment, effectively simulating real-world smart home scenarios.

\subsubsection{Instruction Quality}
In selecting the instruction generation method, we conducted a systematic comparative study. The experimental design encompassed two key dimensions. First, we compared the instruction generation capabilities of several large language models (including ChatGPT-3.5, GPT-4, and Qwen-72B). Second, we established a benchmark control group comprising human-written instructions. For each method, we generated 200 instruction samples and performed a comprehensive evaluation based on three critical metrics: generation quality, generation efficiency, and economic cost.

The experimental results revealed that in terms of generation quality, ChatGPT-3.5 performed comparably to human-written instructions. However, it demonstrated significant advantages in both efficiency and cost-effectiveness. Specifically, ChatGPT-3.5 required an average of only 2 minutes to generate instructions at a cost of less than 0.5 dollars, whereas manual instruction writing took nearly 4 hours with an approximate cost of \$25.

Based on these findings—comparable quality but substantially superior efficiency and cost—we ultimately selected ChatGPT-3.5 as our instruction generation tool.

\subsection{Experiment}
\label{subsec:es}

\textbf{Implementation Details}. Our experiments were conducted using a combination of open-source models, closed-source models, and fine-tuned models, leveraging different hardware and APIs as appropriate. For open-source models, all experiments were performed on NVIDIA RTX 3090 GPUs, ensuring efficient execution and evaluation. For closed-source models and Qwen2.5-72B-Instruct, we utilized APIs provided by OpenAI, enabling us to benchmark their performance under the same experimental conditions. Fine-tuning experiments were conducted on NVIDIA A800 GPUs, leveraging their high memory capacity and computational power to optimize model performance. We employed Hugging Face’s PEFT (Parameter-Efficient Fine-Tuning) framework to fine-tune the models efficiently while minimizing computational costs. As shown in the figure \ref{fig:lora}, we use LoRA to fine-tune the Qwen model.

\begin{figure}
    \centering
    \begin{lstlisting}[
    numbers=left, % 显示行号
    numberstyle=\tiny, % 行号字体
    keywordstyle=\color{blue!70}, % 关键字颜色
    commentstyle=\color{red!50!green!50!blue!50}, % 注释颜色
    frame=shadowbox, % 为代码块添加阴影框
    rulesepcolor=\color{red!20!green!20!blue!20}, % 阴影框颜色
    escapeinside=``, % 允许在代码块中使用 LaTeX 命令
    xleftmargin=2em, xrightmargin=2em, aboveskip=1em, % 设置代码块的边距
    framexleftmargin=2em,% 阴影框左边距
    language=python,
    breaklines=true,
]
lora_config = LoraConfig(
    r=16,
    lora_alpha=32,
    target_modules=["q_proj", "k_proj", "v_proj", "o_proj", "gate_proj", "up_proj", "down_proj"],
    lora_dropout=0.1,
    bias="none",
    task_type="CAUSAL_LM"
)
\end{lstlisting}
    \caption{The LoRA configuration used for fine-tuning the Qwen2.5-7B-Instruct model.}
    \label{fig:lora}
\end{figure}

Table \ref{tabl6} presents the prompt designs used in all model tests. Each prompt contains information about the status of room devices to ensure that the model accurately understands the current environment and generates appropriate responses (see Table \ref{tabl7}). Additionally, we have listed the invocation methods for all devices to enable the model to execute relevant operations correctly, as detailed in Table \ref{tabl8}.

To determine the most suitable prompt, we conducted experiments with four different prompts. Prompt 1 and Prompt 2 were manually written to explore the effectiveness of human-designed prompts. Prompt 3 was derived from Prompt 1 but refined using GPT-4 to evaluate the capability of large language models in prompt optimization. Prompt 4 was also manually written and was ultimately selected as the final prompt version. Figure \ref{tabl10} illustrates the performance of different prompts on the test dataset, comparing their effectiveness and providing insights into the selection of the optimal prompt.

\subsubsection{Few Shot}
\begin{figure}[htb]
\centering
\resizebox{\columnwidth}{!}{
$
\begin{matrix}
\begin{tikzpicture}
\begin{axis}[
legend columns=2,
xlabel={Shot Numbers},
    xmin=0, xmax=5,
    ymin=0, ymax=100,
    xtick={0,1,2,3,4,5},
    ytick={0,20,40,60,80,100},
    title={SUCC},
    legend style={ 
        legend pos =north east,legend columns=-1,}
]
\addplot[
    color=blue,
    mark=square,
    ] coordinates{
    (0,17.07)
    (1,22.87)
    (2,23.22)
    (3,25.00)
    (4,26.93)
    (5,26.18)
    };
    \addlegendentry{ALL}
\addplot[
    color=red,
    mark=star,
    ] coordinates{
    (0,47.11)
    (1,63.36)
    (2,67.33)
    (3,68.35)
    (4,71.39)
    (5,68.98)
    };
    \addlegendentry{VS}
\addplot[
    color=green,
    mark=diamond,
    ] coordinates{
    (0,0.61)
    (1,0.27)
    (2,1.46)
    (3,14.12)
    (4,2.43)
    (5,2.84)
    };
    \addlegendentry{IS}
\addplot[
    color=purple,
    mark=triangle,
    ] coordinates{
    (0,3.27)
    (1,13.88)
    (2,31.02)
    (3,39.18)
    (4,34.29)
    (5,31.84)
    };
    \addlegendentry{VM}
\addplot[
    color=orange,
    mark=+,
    ] coordinates{
    (0,0.0)
    (1,0)
    (2,0)
    (3,0.02)
    (4,0.27)
    (5,0.20)
    };
    \addlegendentry{MM}
\addplot[
    color=yellow,
    mark=*,
    ] coordinates{
    (0,0.0)
    (1,0)
    (2,0)
    (3,0)
    (4,0)
    (5,0)
    };
    \addlegendentry{IM}
\end{axis}
\end{tikzpicture}
&
\begin{tikzpicture}
\begin{axis}[legend columns=2,
xlabel={Shot Numbers},
    xmin=0, xmax=5,
    ymin=0, ymax=100,
    xtick={0,1,2,3,4,5},
    ytick={0,20,40,60,80,100},
    title={F1},
    legend style={ 
        legend pos =north east,legend columns=-1,}
]
\addplot[
    color=blue,
    mark=square,
    ] coordinates{
    (0,18.21)
    (1,25.87)
    (2,26.43)
    (3,29.31)
    (4,32.69)
    (5,31.94)
    };
    \addlegendentry{ALL}
\addplot[
    color=red,
    mark=star,
    ] coordinates{
    (0,47.36)
    (1,63.25)
    (2,62.10)
    (3,67.43)
    (4,71.46)
    (5,69.01)
    };
    \addlegendentry{VS}
\addplot[
    color=green,
    mark=diamond,
    ] coordinates{
    (0,0.61)
    (1,0.28)
    (2,2.39)
    (3,1.45)
    (4,2.56)
    (5,2.81)
    };
    \addlegendentry{IS}
\addplot[
    color=purple,
    mark=triangle,
    ] coordinates{
    (0,31.83)
    (1,44.36)
    (2,38.23)
    (3,59.96)
    (4,62.05)
    (5,61.87)
    };
    \addlegendentry{VM}
\addplot[
    color=orange,
    mark=+,
    ] coordinates{
    (0,2.26)
    (1,23.19)
    (2,23.96)
    (3,26.84)
    (4,30.77)
    (5,30.09)
    };
    \addlegendentry{MM}
\addplot[
    color=yellow,
    mark=*,
    ] coordinates{
    (0,0)
    (1,0)
    (2,2.11)
    (3,2.22)
    (4,2.25)
    (5,3.58)
    };
    \addlegendentry{IM}
\end{axis}
\end{tikzpicture}
\end{matrix}
$
}
\caption{The performance of Llama3-8B-Instruct model in adding different types of data samples (the order of adding is: VS, IS, VM, MM, IM).}
\label{f-fewshot1}
\end{figure}
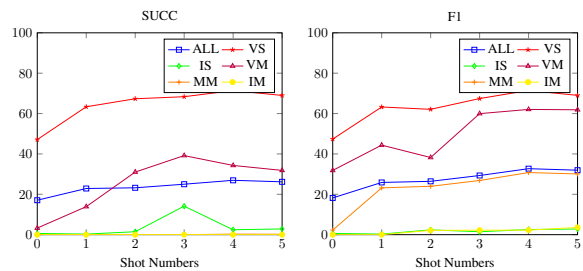

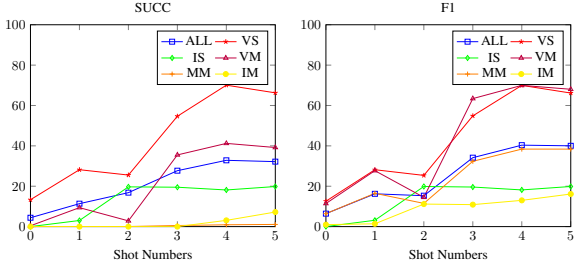
\begin{figure}[htb]
\centering
\resizebox{\columnwidth}{!}{
$
\begin{matrix}
\begin{tikzpicture}
\begin{axis}[
legend columns=2,
xlabel={Shot Numbers},
    xmin=0, xmax=5,
    ymin=0, ymax=100,
    xtick={0,1,2,3,4,5},
    ytick={0,20,40,60,80,100},
    title={SUCC},
    legend style={ 
        legend pos =north east,legend columns=-1,}
]
\addplot[
    color=blue,
    mark=square,
    ] coordinates{
    (0,4.39)
    (1,11.36)
    (2,16.82)
    (3,27.703)
    (4,32.845)
    (5,32.189)
    };
    \addlegendentry{ALL}
\addplot[
    color=red,
    mark=star,
    ] coordinates{
    (0,13.32)
    (1,28.19)
    (2,25.537)
    (3,54.66)
    (4,70.066)
    (5,66.267)
    };
    \addlegendentry{VS}
\addplot[
    color=green,
    mark=diamond,
    ] coordinates{
    (0,0)
    (1,3.03)
    (2,19.689)
    (3,19.512)
    (4,18.15)
    (5,19.85)
    };
    \addlegendentry{IS}
\addplot[
    color=purple,
    mark=triangle,
    ] coordinates{
    (0,0.41)
    (1,9.39)
    (2,2.85)
    (3,35.51)
    (4,41.224)
    (5,39.1836)
    };
    \addlegendentry{VM}
\addplot[
    color=orange,
    mark=+,
    ] coordinates{
    (0,0.0)
    (1,0.02)
    (2,0.049)
    (3,0.54)
    (4,0.908)
    (5,1.08)
    };
    \addlegendentry{MM}
\addplot[
    color=yellow,
    mark=*,
    ] coordinates{
    (0,0.0)
    (1,0)
    (2,0)
    (3,0)
    (4,3.09)
    (5,7.216)
    };
    \addlegendentry{IM}
\end{axis}
\end{tikzpicture}
&
\begin{tikzpicture}
\begin{axis}[legend columns=2,
xlabel={Shot Numbers},
    xmin=0, xmax=5,
    ymin=0, ymax=100,
    xtick={0,1,2,3,4,5},
    ytick={0,20,40,60,80,100},
    title={F1},
    legend style={ 
        legend pos =north east,legend columns=-1,}
]
\addplot[
    color=blue,
    mark=square,
    ] coordinates{
    (0,6.42)
    (1,16.24)
    (2,15.313)
    (3,34.148)
    (4,40.35)
    (5,40.023)
    };
    \addlegendentry{ALL}
\addplot[
    color=red,
    mark=star,
    ] coordinates{
    (0,12.58)
    (1,28.18)
    (2,25.37)
    (3,54.86)
    (4,69.97)
    (5,66.17)
    };
    \addlegendentry{VS}
\addplot[
    color=green,
    mark=diamond,
    ] coordinates{
    (0,0.02)
    (1,3.14)
    (2,19.8778)
    (3,19.602)
    (4,18.1737)
    (5,19.8727)
    };
    \addlegendentry{IS}
\addplot[
    color=purple,
    mark=triangle,
    ] coordinates{
    (0,11.44)
    (1,27.71)
    (2,14.499)
    (3,63.417)
    (4,69.965)
    (5,68.04)
    };
    \addlegendentry{VM}
\addplot[
    color=orange,
    mark=+,
    ] coordinates{
    (0,6.4)
    (1,16.49)
    (2,11.508)
    (3,32.36)
    (4,38.3932)
    (5,38.39)
    };
    \addlegendentry{MM}
\addplot[
    color=yellow,
    mark=*,
    ] coordinates{
    (0,0.98)
    (1,1.395)
    (2,11.165)
    (3,10.884)
    (4,13.004)
    (5,16.143)
    };
    \addlegendentry{IM}
\end{axis}
\end{tikzpicture}
\end{matrix}
$
}
\caption{The performance of Gemma2-9B-Instruct model in adding different types of data samples (the order of adding is: VS, IS, VM, MM, IM).}
\label{f-fewshot2}
\end{figure}

\begin{figure}[htb]
\centering
\resizebox{\columnwidth}{!}{
$
\begin{matrix}
\begin{tikzpicture}
\begin{axis}[
legend columns=2,
xlabel={Shot Numbers},
    xmin=0, xmax=5,
    ymin=0, ymax=100,
    xtick={0,1,2,3,4,5},
    ytick={0,20,40,60,80,100},
    title={SUCC},
    legend style={ 
        legend pos =north east,legend columns=-1,}
]
\addplot[
    color=blue,
    mark=square,
    ] coordinates{
    (0,0.9)
    (1,21.219)
    (2,17.3096)
    (3,15.674)
    (4,12.726)
    (5,13.15)
    };
    \addlegendentry{ALL}
\addplot[
    color=red,
    mark=star,
    ] coordinates{
    (0,2.49)
    (1,59.46)
    (2,47.923)
    (3,42.686)
    (4,34.572)
    (5,35.849)
    };
    \addlegendentry{VS}
\addplot[
    color=green,
    mark=diamond,
    ] coordinates{
    (0,0.01)
    (1,0)
    (2,0.576)
    (3,0.458)
    (4,0.339)
    (5,0.3547)
    };
    \addlegendentry{IS}
\addplot[
    color=purple,
    mark=triangle,
    ] coordinates{
    (0,0.82)
    (1,2.448)
    (2,0.408)
    (3,20.408)
    (4,19.59)
    (5,17.142)
    };
    \addlegendentry{VM}
\addplot[
    color=orange,
    mark=+,
    ] coordinates{
    (0,0.0)
    (1,0)
    (2,0.02)
    (3,0.0)
    (4,0)
    (5,0)
    };
    \addlegendentry{MM}
\addplot[
    color=yellow,
    mark=*,
    ] coordinates{
    (0,0.0)
    (1,0)
    (2,0)
    (3,0)
    (4,0)
    (5,0)
    };
    \addlegendentry{IM}
\end{axis}
\end{tikzpicture}
&
\begin{tikzpicture}
\begin{axis}[legend columns=2,
xlabel={Shot Numbers},
    xmin=0, xmax=5,
    ymin=0, ymax=100,
    xtick={0,1,2,3,4,5},
    ytick={0,20,40,60,80,100},
    title={F1},
    legend style={ 
        legend pos =north east,legend columns=-1,}
]
\addplot[
    color=blue,
    mark=square,
    ] coordinates{
    (0,2.33)
    (1,19.49)
    (2,12.19)
    (3,23.923)
    (4,23.574)
    (5,24.1145)
    };
    \addlegendentry{ALL}
\addplot[
    color=red,
    mark=star,
    ] coordinates{
    (0,4.68)
    (1,59.902)
    (2,53.168)
    (3,50.07)
    (4,44.02)
    (5,44.638)
    };
    \addlegendentry{VS}
\addplot[
    color=green,
    mark=diamond,
    ] coordinates{
    (0,0.03)
    (1,0.01)
    (2,0.706)
    (3,0.5688)
    (4,0.4458)
    (5,0.488)
    };
    \addlegendentry{IS}
\addplot[
    color=purple,
    mark=triangle,
    ] coordinates{
    (0,6.23)
    (1,18.3469)
    (2,2.05)
    (3,53.53)
    (4,51.603)
    (5,50.586)
    };
    \addlegendentry{VM}
\addplot[
    color=orange,
    mark=+,
    ] coordinates{
    (0,2.26)
    (1,23.19)
    (2,1.677)
    (3,22.987)
    (4,23.8169)
    (5,24.539)
    };
    \addlegendentry{MM}
\addplot[
    color=yellow,
    mark=*,
    ] coordinates{
    (0,0)
    (1,0)
    (2,0.624)
    (3,0)
    (4,0)
    (5,0.49)
    };
    \addlegendentry{IM}
\end{axis}
\end{tikzpicture}
\end{matrix}
$
}
\caption{The performance of Mistral-7B-v0.3 model in adding different types of data samples (the order of adding is: VS, IS, VM, MM, IM).}
\label{f-fewshot3}
\end{figure}
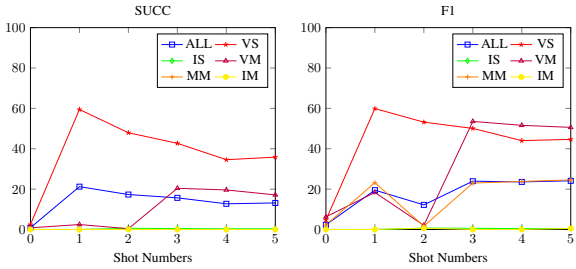

\begin{table*}[]
\centering
\begin{tcolorbox}[colback=gray!10,
			colframe=black,
			width=14cm,
			arc=2mm, auto outer arc,
			title={Few shot},	]
Here are a few examples, your output format should be consistent with the results provided in the example:\\
<example1>

    \quad<User instructions:> Lower the air conditioner temperature in the guest bedroom by 1 degree.  
    
    \quad<Machine instructions:> \{guest\_bedroom.air\_conditione.set\_temperature(26)\}    
    
</example1>

<example2>    

    \quad<User instructions:> Turn on the light in the living room.
    
    \quad<Machine instructions:> \{error\_input\}  (If there is no device or if there is no attribute or method that can be operated on the device, please output as "error\_input".)    
    
</example2>

<example3>

    \quad<User instructions:> Set the intensity of the humidifier to 20\% in the guest bedroom, adjust the intensity of the dehumidifiers by increasing it by 30\% in the study room, and set the intensity of the dehumidifiers to 40\% in the study room.    
    
    \quad<Machine instructions:>\{guest\_bedroom.humidifier.set\_intensity(20),study\_room.de
    humidifiers.set\_intensity(10),study\_room.dehumidifiers.set\_intensity(40),\}
    
</example3>

<example4>

    \quad<User instructions:> Increase the intensity of the aromatherapy in the corridor by 4\%, set the brightness of the lights in the garage to 40, adjust the fan speed of the air conditioner in the living room to low, and set the volume of the media player in the living room to 0.   
    
    \quad<Machine instructions:> \{corridor.aromatherapy.set\_intensity(30),error\_input,error\_
    input,living\_room.media\_player.set\_volume(0),\}   
    
</example4>

<example5>

    \quad<User instructions:> Turn on the air purifiers in the garage and increase the volume of the media player in the living room by 50 percent.    
    
    \quad<Machine instructions:> \{error\_input,error\_input\}
    
</example5>

\end{tcolorbox}	
    \caption{Different types of few-shot, the sequence from top to bottom is VS, IS, VM, MM, IM.}
    \label{tabl9}
\end{table*}

Table \ref{tabl9} presents the different categories of samples used to evaluate various few-shot learning methods. These samples cover a range of device function categories, listed from top to bottom: VS, IS, VM, MM, and ME. By incorporating these diverse categories, we can more effectively assess the model’s performance across different real-world smart home scenarios.

Figures \ref{f-fewshot1}, \ref{f-fewshot2}, and \ref{f-fewshot3} illustrate the performance of Llama3-8B-Instruct, Gemma2-9B-Instruct, and Mistral-7B-v0.3, respectively, under different few-shot settings. The results from Llama3-8B-Instruct and Gemma2-9B-Instruct align with our findings in Section \ref{few-shot}: as the number of few-shot examples increases, the models generally exhibit improved performance, though the degree of enhancement varies across different data types. However, we also observed a performance decline in certain tasks, particularly IS and VS, as the number of shots increased. This decline is likely due to the lack of home device state information in the additional shots. Since the added examples only include user instructions and device operations, the model struggles to accurately determine device status, leading to confusion and a drop in performance.

Interestingly, Mistral-7B-v0.3 displays a unique behavior: it reaches its performance ceiling at just 2-shot, with no significant improvement beyond this point. This could be attributed to the fact that we did not use an instruction-tuned version of Mistral, making it less capable of fully understanding the specific requirements of our task. This also explains Mistral’s relatively poor performance in the main experiments, further reinforcing our hypothesis.

\subsubsection{RAG Setting and Case Study}
We segmented device states and methods based on room categories and input these segments into the LLM to extract hidden states from the final layer. This approach allowed us to capture meaningful representations of device functionality and contextual dependencies.

Following the methodology outlined in \citep{10.1007/978-981-97-5669-8_5}, we incorporated the phrase "After thinking step by step" into the prompt. This addition helps the model better process sequential reasoning and compress semantic information effectively, improving its understanding of device interactions. After obtaining embeddings for both the segmented chunks and user instructions, we computed cosine similarity to facilitate retrieval. The retrieval process follows these conditions:
\begin{itemize}
    \item If the similarity score exceeds 0.5 and the number of retrieved chunks is greater than 3, we return the results directly.
    \item If fewer than 3 chunks are retrieved, we return the top three chunks with the highest similarity scores to ensure the model has sufficient contextual information for accurate execution.
\end{itemize}

This strategy enhances retrieval efficiency and ensures that the model has a reliable and relevant device context for decision-making.

\begin{table*}[!htb]
\resizebox{\linewidth}{!}{%
\begin{tabular}{c|c|c|c}
\hline
Data Type & Input                                                                                                                                                                                                                                                                                                                                                       & Golden                                                                                                                                                                                                 & Genrated                                                                                                              \\ \hline
VS        & Set the brightness of the light on the balcony to 50. balcony.light.set\_brightness(50)                                                                                                                                                                                                                                                                     & balcony.light.set\_brightness(50)                                                                                                                                                                      & \textcolor{BrickRed}{error\_input}                                                                                                          \\ \hline
IS        & Pause the media player in the living room.                                                                                                                                                                                                                                                                                                                  & error\_input                                                                                                                                                                                           & \textcolor{OliveGreen}{error\_input}                                                                                                          \\ \hline
VM        & \begin{tabular}[c]{@{}c@{}}Lower the air conditioner temperature in the guest bedroom to 20 degrees,\\ set the brightness of the light in the foyer to 50,\\ and increase the intensity of the dehumidifier in the study room by 32\%.\end{tabular}                                                                                                         & \begin{tabular}[c]{@{}c@{}}guest\_bedroom.air\_conditioner.set\_temperature(20),\\ foyer.light.set\_brightness(50),\\ study\_room.dehumidifiers.set\_intensity(70)\end{tabular}                        & \begin{tabular}[c]{@{}c@{}}\textcolor{BrickRed}{error\_input},\\ \textcolor{BrickRed}{error\_input},\\ \textcolor{BrickRed}{error\_input,}\end{tabular}                                 \\ \hline
IM        & \begin{tabular}[c]{@{}c@{}}Set the intensity of the humidifier to 60 in the study room,\\ adjust the volume of the media player to 60 on the balcony,\\ and set the degree of the curtain to 20 on the balcony.\end{tabular}                                                                                                                                & \begin{tabular}[c]{@{}c@{}}error\_input,\\ error\_input,\\ error\_input,\end{tabular}                                                                                                                  & \begin{tabular}[c]{@{}c@{}}\textcolor{OliveGreen}{error\_input,}\\ \textcolor{OliveGreen}{error\_input,}\\ \textcolor{OliveGreen}{error\_input,}\end{tabular}                                 \\ \hline
MM        & \begin{tabular}[c]{@{}c@{}}Set the air conditioner temperature to 16 degrees in the guest bedroom,\\ turn off the lights in the study room,\\ decrease the media player volume by 30 percent on the balcony,\\ set the dehumidifier intensity to 60 in the master bedroom,\\ and adjust the heating temperature to 27 degrees in the bathroom.\end{tabular} & \begin{tabular}[c]{@{}c@{}}guest\_bedroom.air\_conditioner.set\_temperature(16),\\ study\_room.light.turn\_off(),\\ error\_input,\\ error\_input,\\ bathroom.heating.set\_temperature(27)\end{tabular} & \begin{tabular}[c]{@{}c@{}}\textcolor{BrickRed}{error\_input},\\ \textcolor{BrickRed}{error\_input},\\ \textcolor{OliveGreen}{error\_input,}\\ \textcolor{OliveGreen}{error\_input,}\\ \textcolor{BrickRed}{error\_input},\end{tabular} \\ \hline
\end{tabular}%
}
\caption{Model output when there no useful information in the context.
\textcolor{BrickRed}{Red} indicates an incorrect output, while \textcolor{OliveGreen}{green} represents a correct output.}
\label{rag case}
\end{table*}

In the Table \ref{rag case}, we provide case examples of context retrieval errors across different data categories. This context mismatch leads to a counterintuitive phenomenon—performance degradation on simple and effective tasks, but unexpected improvement on more challenging ones. Upon further analysis, we find that the root cause lies in insufficient embedding similarity between device states/methods and user instructions, which hampers the similarity-based retrieval mechanism’s ability to effectively distinguish between different context fragments. To address this issue, we propose training a specialized retrieval optimization model to enhance context alignment accuracy.

\subsubsection{Error Analysis}
Table \ref{error anslys type} provides a detailed breakdown of the proportion and quantity of different error types across various data types. It presents the distribution of Unfaithfulness, In-context Attention Error, and Key Error for each data type, along with their respective percentages and absolute counts. This allows for a more intuitive analysis of the error patterns and potential issues affecting the model's performance across different data types.

\begin{table}[t]
\centering
\resizebox{\columnwidth}{!}{%
\begin{tabular}{c|ccc}
\hline
Type & Unfaithfulness & In-context Attention Error & Key Error  \\ \hline
VS   & 50.53/805      & 7.53/120                   & 41.93/668  \\
IS   & -              & 100/873                    & -          \\
VM   & 53.83/71       & 60.90/81                   & 44.36/59   \\
IM   & -              & 100/37                     & -          \\
MM   & 23.45/732      & 44.44/1387                 & 41.15/1253 \\ \hline
\end{tabular}%
}
\caption{The proportion of different error types for different data types. The former is the ratio, and the latter is the quantity.}
\label{error anslys type}
\end{table}

\subsubsection{OOD}
\begin{figure}[t]
    \centering
    \resizebox{\columnwidth}{!}{
    \begin{tikzpicture}
    \begin{axis}
    [ybar , 
    grid=major,major grid style={dashed}, 
    ymin=0,  
    bar width=1cm, 
    ylabel=Nums,  
    width=10cm,
    height=6cm,  
    symbolic x coords={VS,IS,VM,MM,IM,ALL},  
    x=1.5cm, 
    xtick=data, 
    nodes near coords, 
    nodes near coords style={font=\fontsize{8}{12}\selectfont}, 
    enlarge x limits=0.2, 
    legend style={at={(0.5,-0.2)},anchor=north,legend columns=-1}, 
    ] 
    \addplot[fill=CornflowerBlue,color=CornflowerBlue] coordinates {(VS,565)(IS,502)(VM,44)(MM,106)(IM,90)(ALL,1207)};
    \end{axis} 
    \end{tikzpicture}
    }
    \caption{OOD data distribution.}
\label{ood}
\end{figure}
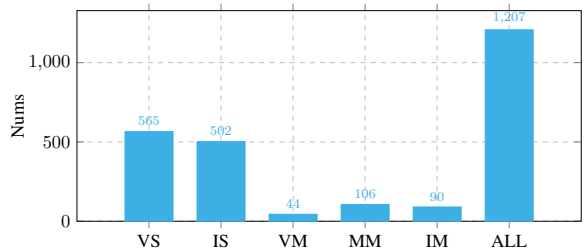

We have created two new smart devices: beds and pet feeders to construct ood data. The data distribution is shown in the Figure \ref{ood}.

\begin{table}[]
\resizebox{\columnwidth}{!}{%
\begin{tabular}{c|c|c|c|l|l|l}
\hline
Models & LLaMa3-8b & Qwen2.5-7B & Mistral-7B & Gemma2-9B & Deepseek-V3 & GPT-4o \\ \hline
Latecy & 2.49s      & 1.99s       & 3.05s       & 6.92s      & 6.79s        & 1.16s   \\ \hline
\end{tabular}%
}
\caption{Inferface latecy}
\label{tab:il}
\end{table}

\subsubsection{Inference Latency}
In Table \ref{tab:il}, our tests conducted on an RTX 3090 platform demonstrate that the inference latency of the locally deployed model can be kept within a few seconds, which generally satisfies practical requirements. However, to further improve the user experience—ideally by reducing latency to under one second—enhancing inference speed remains a key focus of our future work.
\end{document}